\documentclass{article}
\usepackage{ijcai25}

\usepackage{times}
\usepackage{soul}
\usepackage{url}
\usepackage[hidelinks]{hyperref}
\usepackage[utf8]{inputenc}
\usepackage[small]{caption}
\usepackage{graphicx}
\usepackage{amsmath}
\usepackage{amsthm}
\usepackage{booktabs}
\usepackage{algorithm}
\usepackage{algorithmic}
\usepackage[switch]{lineno}

\usepackage{amsmath}
\usepackage{amssymb}
\usepackage{bm} 
\usepackage{makecell} 
\usepackage{multirow} 

\title{Adaptive Language-Aware Image Reflection Removal Network}

\author{
	Siyan Fang$^1$\and
	Yuntao Wang$^1$\and
	Jinpu Zhang$^2$\and
	Ziwen Li$^1$\And
	Yuehuan Wang$^1$\thanks{Corresponding author.}
	\\
	\affiliations
	$^1$Huazhong University of Science and Technology\\
	$^2$National University of Defense Technology\\
	\emails
	fangsiyanfsy@163.com,
	yuntaowang@hust.edu.cn,
	zhangjinpu@nudt.edu.cn,
	\{D201980722, yuehwang\}@hust.edu.cn
}

\begin{document}

\maketitle

\begin{abstract}
Existing image reflection removal methods struggle to handle complex reflections. Accurate language descriptions can help the model understand the image content to remove complex reflections. However, due to blurred and distorted interferences in reflected images,  machine-generated language descriptions of the image content are often inaccurate, which harms the performance of language-guided reflection removal.  To address this, we propose the Adaptive Language-Aware Network (ALANet) to remove reflections even with inaccurate language inputs. Specifically, ALANet integrates both filtering and optimization strategies. The filtering strategy reduces the negative effects of language while preserving its benefits, whereas the optimization strategy enhances the alignment between language and visual features. ALANet also utilizes language cues to decouple specific layer content from feature maps, improving its ability to handle complex reflections. To evaluate the model's performance under complex reflections and varying levels of language accuracy, we introduce the Complex Reflection and Language Accuracy Variance (CRLAV) dataset. Experimental results demonstrate that ALANet surpasses state-of-the-art methods for image reflection removal. The code and dataset are available at \href{https://github.com/fashyon/ALANet}{https://github.com/fashyon/ALANet}.
\end{abstract}

\section{Introduction}

When capturing images through glass, reflections diminish image quality by obscuring details and  distorting colors, thereby impairing the image's usability and hindering downstream computer vision tasks \cite{xie2021segformer,zhang2023progressive}. The widely used reflection model \cite{wan2018crrn,yang2018seeing} considers the reflected image $\bm{I}$ as a linear combination of the transmission layer $\bm{T}$ and the reflection layer $\bm{R}$, where $\bm{T}$ represents the content transmitted through the glass, and $\bm{R}$ represents the reflected content.
\begin{figure}[t]
	\centering
	\includegraphics[width=0.98\columnwidth]{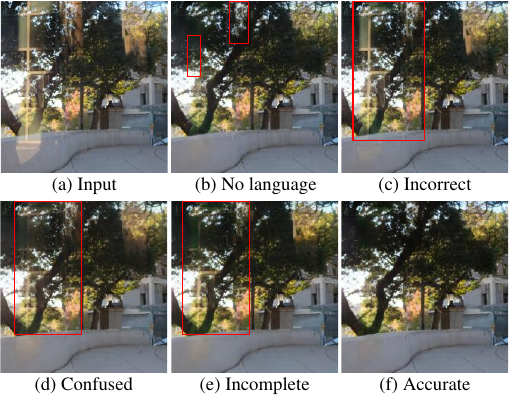} 
	\caption{The impact of language-guided reflection removal with different types of language inputs. Inaccurate language inputs result in worse outcomes than having no language. The specific language inputs for each subfigure are provided in the supplementary material.}
	\label{fig1_1}
\end{figure}

To obtain a clear $\bm{T}$ from $\bm{I}$, multi-image reflection removal methods \cite{li2013exploiting,liu2020learning} leverage the distinct motions between  $\bm{T}$ and $\bm{R}$ from different viewpoints to distinguish them. However, these methods rely on controlled environments and multi-angle observations, limiting their applicability in real-world scenarios. Earlier single-image reflection removal methods \cite{levin2004separating,chung2009interference,shih2015reflection} primarily rely on handcrafted priors, but are only suitable for simple reflection scenarios. With the rise of deep learning, many methods \cite{zhang2018single,wei2019single,li2020single,hu2023single}  use neural networks to model the different layers in reflected images. However, these methods struggle to remove complex reflections due to the limited information  available in a single image.

Language has shown great effectiveness in various visual tasks \cite{radford2021learning,li2022blip,wang2022cris}. Language descriptions can provide additional information about objects within a scene, enhancing the understanding and processing capabilities of networks. For example,  with complex reflections in an image, language descriptions can indicate which areas belong to the reflection layer and which belong to the transmission layer. Zhong et al.  \cite{zhong2024language} introduced language into the image reflection removal field to provide contextual clues, which facilitate the separation of different layers. However, this method requires that the language descriptions accurately match the image content; mismatches can harm performance, as illustrated in Figure \ref{fig1_1}. The L-DiffER \cite{hong2025differ} employed a language-based diffusion model to remove reflections, but does not account for cases where the language descriptions are inaccurate. Since manually annotating images is time-consuming and labor-intensive, the most convenient approach is to use language models to automatically generate descriptions. However, reflections interfere with the language model's ability to understand the image content. For example, on the reflection removal datasets Nature, Real, and SIR\textsuperscript{2}, using BLIP \cite{li2022blip} to generate captions for reflected images  versus reflection-free images, the average  BLEU, METEOR, CIDEr, and ROUGE-L scores drop by 50.95\%, 34.11\%, 55.07\%, and 31.45\%, respectively, indicating that reflections mislead the language model into generating inaccurate descriptions. These inaccuracies fall into three categories: (1) Incorrect: Describing non-existent content in the image, misleading the model to process fictitious elements and adversely affecting its handling of actual content. (2) Confused: Mixing up of parts of the transmission and reflection layers, preventing the model from distinguishing the features of each layer. (3) Incomplete: Omitting details, especially those obscured by reflections, which prevents the model from focusing on the critical areas that need processing. 

To mitigate the negative impact of inaccurate language descriptions on the performance of image reflection removal, we propose the Adaptive Language-Aware Network (ALANet) through two well-designed strategies: filtering and optimization. The filtering strategy aims to filter out the negative effects of inaccurate language while retaining its positive effects. For this, we propose the Language-Aware Competition Attention Module (LCAM), which enables language-guided attention and visually-driven attention to compete with each other, dynamically adjusting their influence. The optimization strategy seeks to refine language features so they can align with the content of the corresponding layers.  For this, we propose the Adaptive Language Calibration Module (ALCM), which uses visual features to fine-tune language features. Furthermore, to effectively utilize language features for separating specific information from images, we design the Language-Guided Spatial-Channel Cross Attention (LSCA) mechanism. This mechanism adjusts the spatial and channel structures of the feature map by language, accurately extracting different layers from intertwined scenes. 

To evaluate model performance under complex reflections and varying levels of language accuracy, we introduce the real-world Complex Reflection and Language Accuracy Variance (CRLAV) dataset. This dataset includes multiple types of complex reflections, with each image paired with language descriptions of varying accuracies. It not only allows for assessing a model's ability to remove complex reflections, but it also evaluates the model's robustness under conditions of varying language accuracy guidance.

In summary, the main contributions are as follows:

\begin{itemize}
	\item We propose ALANet to improve the model's reflection removal performance with inaccurate language descriptions through filtering and optimization strategies.
	\item We introduce the real-world CRLAV dataset to evaluate the performance of models under complex reflections and varying levels of language accuracy.
	\item Experiments demonstrate that the proposed ALANet surpasses state-of-the-art (SOTA) methods and achieves solid performance even with inaccurate language inputs.
\end{itemize}

\section{Related Work}

\subsection{Image Reflection Removal}

The existing methods for reflection removal can be divided into multi-image-based and single-image-based methods. While multi-image-based methods leverage information from multiple images to remove reflections, their applicability is limited by data acquisition constraints. This paper focuses on the more broadly applicable single-image reflection removal. Early single-image reflection removal methods primarily relied on priors such as gradient sparsity \cite{levin2004separating,levin2007user}, layer smoothness \cite{chung2009interference,yan2014separation}, and ghosting cues \cite{shih2015reflection}. However, these methods often perform poorly in real-world scenes due to their heavy dependence on prior assumptions.

With the rise of deep learning, CEILNet \cite{fan2017generic} was the first to apply deep learning to the task of single-image reflection removal, using edge features as auxiliary information to eliminate reflections. Zhu et al. \cite{zhu2024revisiting} designed the max-min reflection filter to characterize the reflection locations in paired images. However, these methods often face bottlenecks in complex scenes due to the lack of additional information. Given that language conveys human prior knowledge about the world \cite{deng2023nerdi}, Zhong et al. \cite{zhong2024language} facilitated layer separation by establishing correspondences between language descriptions and image layers. However, when the language descriptions do not accurately match the corresponding layer content, they can have negative effects. L-DiffER \cite{hong2025differ} used language as a condition for the diffusion model to remove reflections, but it did not consider the harm caused by inaccurate language descriptions. To address this issue, we propose the ALANet to reduce the negative impact of inaccurate language descriptions on reflection removal.
\begin{figure*}[t]
	\centering
	\includegraphics[width=1.0\textwidth]{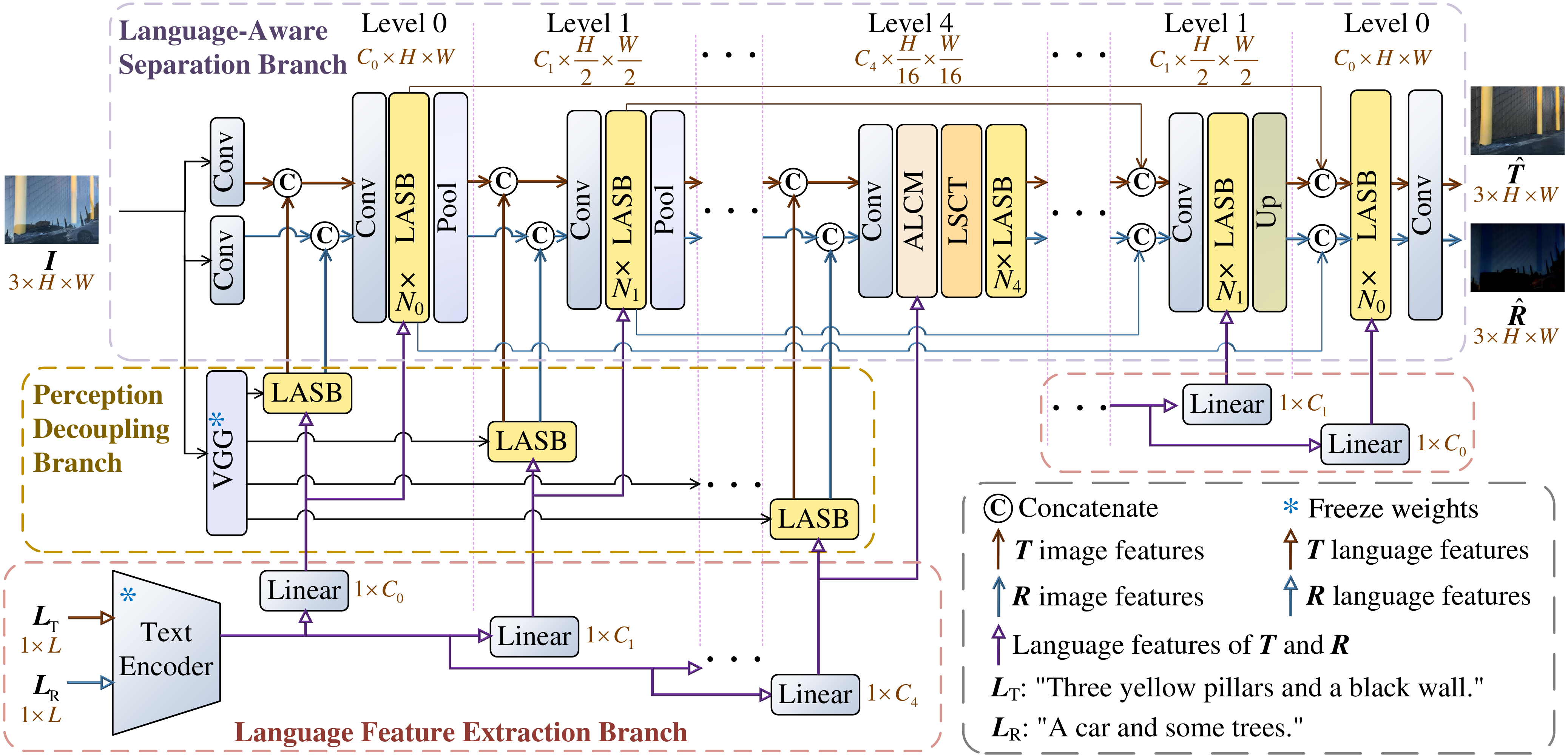} 
	\caption{Overview of the proposed ALANet, which comprises various modules that use language adaptively to remove reflections.  \textbf{\textit{T}} and \textbf{\textit{R}} represent the transmission and reflection layers, respectively.}
	\label{fig2}
\end{figure*}
\subsection{Applications of Language in Image Processing}
The ability of CLIP \cite{radford2021learning} to understand language and interpret images has been leveraged in various tasks. BLIP \cite{li2022blip} proposed using bootstrapping to synthesize captions, obtaining higher quality data. CLIP-LIT \cite{liang2023iterative} learned initial prompt pairs by constraining the text-image similarity between prompts and corresponding images in the CLIP latent space. NeRCo \cite{yang2023implicit} employed the priors of a pre-trained vision-language model to provide perceptually guided instruction for learning lighting conditions. DA-CLIP \cite{luo2023controlling} predicted the degraded embeddings of low-quality images through a controller, and directed the CLIP image encoder to output high-quality content embeddings. Zhong et al. \cite{zhong2024language} introduced language descriptions to provide layer content for addressing the ill-posed problem of reflection separation. However, since images with reflections are difficult to be accurately described by language models, how to separate specific image layers under imperfect language conditions remains an unresolved challenge.
\section{Proposed Method}
\subsection{Adaptive Language-Aware Network}
The Adaptive Language-Aware Network (ALANet) consists of the Language-Aware Separation Branch (LSBranch), the Perception Decoupling Branch (PDBranch), and the Language Feature Extraction Branch (LEBranch), as shown in Figure \ref{fig2}. The LEBranch is responsible for encoding the input language, and adjusting the channel dimensions to fit the network structure. The PDBranch uses a pre-trained Visual Geometry Group (VGG) model \cite{simonyan2014very} to extract high-level visual features. It then decouples the corresponding features through language to facilitate the separation process in the LSBranch. In the LSBranch, the Language-Aware Separation Block (LASB) adjusts the influence of language-guided attention according to the accuracy of the language description, thereby preventing issues of misguidance caused by inaccurate language.
\begin{figure}[t]
	\centering
	\includegraphics[width=1.0\columnwidth]{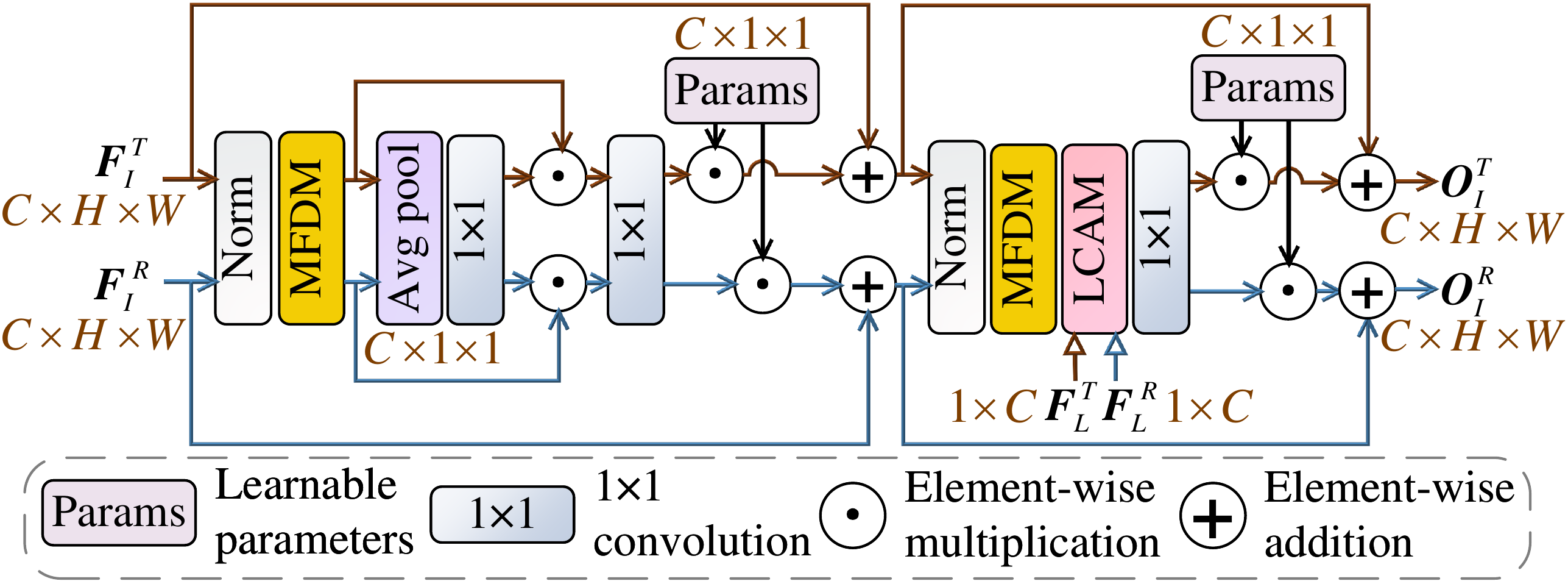} 
	\caption{Structure of the LASB. As the core of LASB, LCAM utilizes language features from different layers to facilitate the separation of those layers.}
	\label{fig3}
\end{figure}

The structure of the LASB is shown in Figure \ref{fig3}. Its purpose is to use semantic information from language that matches with the visual content, guiding the separation of the transmission and reflection layers. The LASB primarily employs the LCAM and the Multi-Receptive Field Decoupling Module (MFDM) to separate the different layers. Details of MFDM are provided in the supplementary material.
\begin{figure}[t]
	\centering
	\includegraphics[width=1.0\columnwidth]{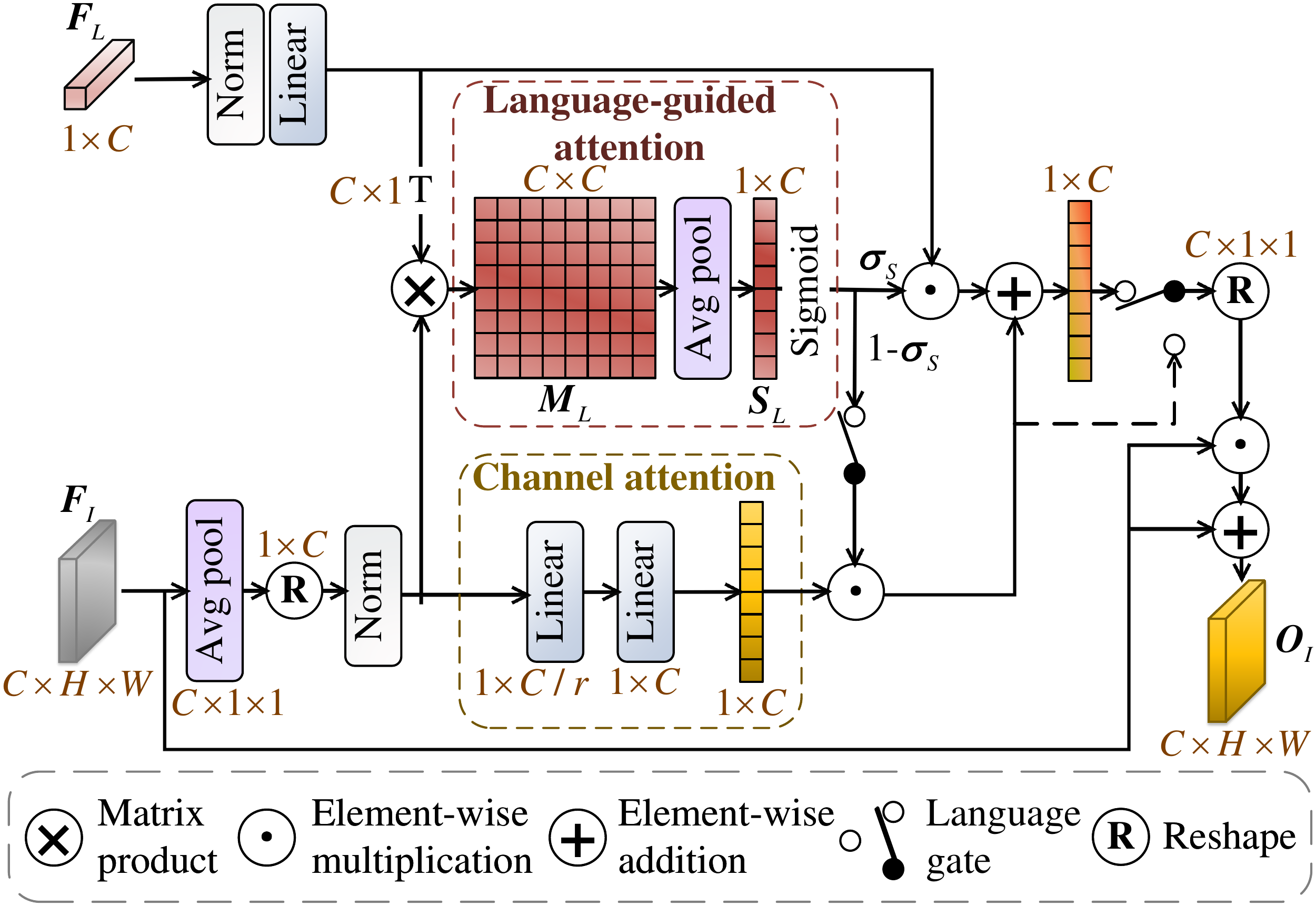} 
	\caption{Structure of the LCAM. The dashed lines indicate the scenario without language input.}
	\label{fig4}
\end{figure}
\subsection{Language-Aware Competition Attention Module}
The Language-Aware Competition Attention Module (LCAM) is shown in Figure \ref{fig4}. It competitively adjusts the weight distribution of language-guided attention and channel attention, based on the matching degree between language cues and the visual content, aiming to retain the positive effects of accurate language while suppressing the negative effects of inaccurate language. Given the image features $\bm{F}_I \in \mathbb{R}^{C \times H \times W}$ and language features $\bm{F}_L \in \mathbb{R}^{1 \times C}$ , where \textit{H} and \textit{W} are the height and width of the image, and $ \textit{C} $ is the number of channels. $\bm{F}_I$ and $\bm{F}_L$ both undergo processing and through matrix multiplication, the language-image similarity matrix  $\bm{M}_L \in \mathbb{R}^{C \times C}$ is obtained. Each element $(i,j)$ in $\bm{M}_L$ represents the similarity between the $i$-th channel of language features and the $j$-th channel of image features. Subsequently,  $\bm{M}_L$ is average-pooled along the language dimension to obtain the language-image similarity scores $\bm{S}_L \in \mathbb{R}^{1 \times C}$, where each element indicates the similarity of each image channel to the overall language information. After passing through a sigmoid function, the adjustment vector $\bm{\sigma}_S$ is used to modify the weight of language-guided attention, and its complement $1 - \bm{\sigma}_S$  is used to adjust the weight of channel attention. The channel attention differentiates between the two layers by considering their physical characteristics, specifically the structural continuity in the transmission layer and the specular sparsity in the reflection layer, thereby complementing language-guided attention from a visual perspective.
\subsection{Adaptive Language Calibration Module}
\begin{figure}[t]
	\centering
	\includegraphics[width=0.95\columnwidth]{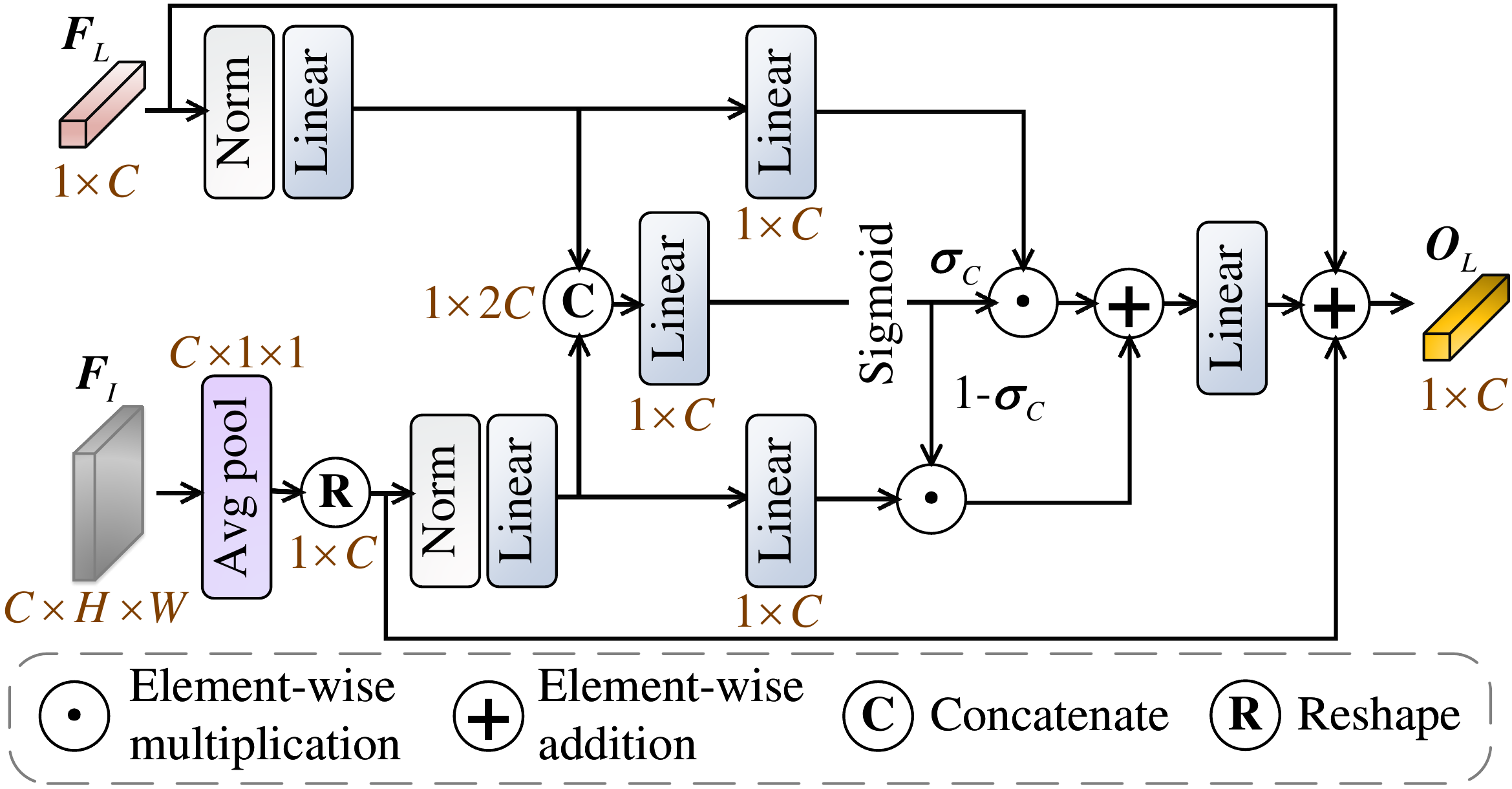} 
	\caption{Structure of the ALCM. The ALCM enhances the consistency between language features and visual content.}
	\label{fig5}
\end{figure}

The Adaptive Language Calibration Module (ALCM), as shown in Figure \ref{fig5}, adjusts and optimizes language features by leveraging visual features to enhance the consistency between language and visual content. In the ALCM, image features $\bm{F}_I$ and language features $\bm{F}_L$ are processed and then concatenated. Subsequently, they pass through a linear layer and a sigmoid function to generate an adjustment vector $\bm{\sigma}_c \in \mathbb{R}^{1 \times C}$, which dynamically controls the fusion ratio of language and image features. In this process, the linear layer acts as a mediator for feature fusion, optimizing the combination points between the two types of information.

\subsection{Language-Guided Spatial-Channel Cross Transformer}
\begin{figure}[t]
	\centering
	\includegraphics[width=1.0\columnwidth]{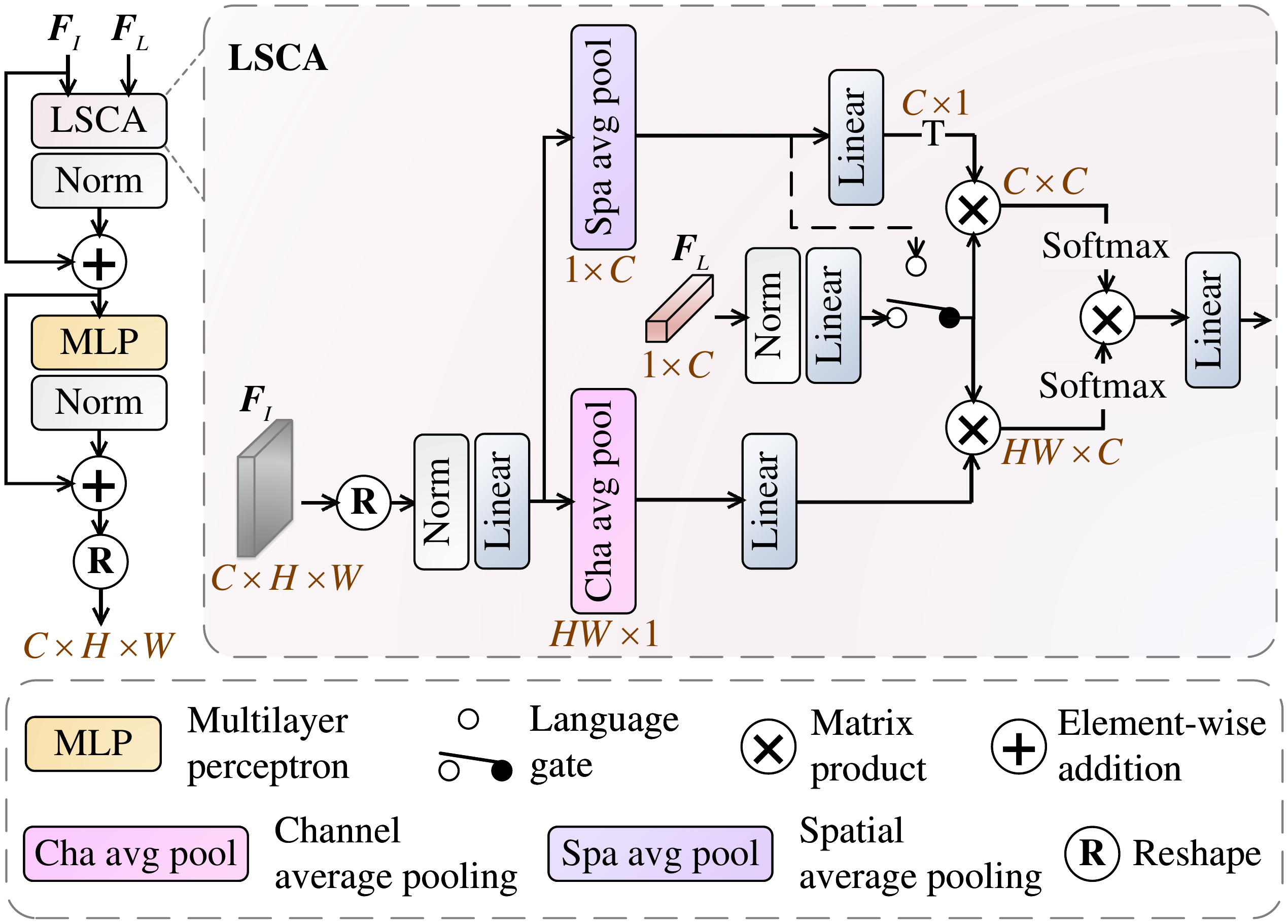} 
	\caption{Structure of the LSCT and its core component, LSCA. The dashed lines in LSCA indicate the scenario without language input.}
	\label{fig6}
\end{figure}
\begin{table*}[t]
	\centering
	\setlength{\tabcolsep}{7pt}
	\resizebox{\textwidth}{!}{
		\begin{tabular}{c|cc|cc|cc|cc|cc|cc}
			\hline
			\multirow{2}{*}{\centering \textbf{Methods}} & \multicolumn{2}{c|}{\textbf{Nature (20)}} & \multicolumn{2}{c|}{\textbf{Real (20)}} & \multicolumn{2}{c|}{\textbf{Wild (55)}} & \multicolumn{2}{c|}{\textbf{Postcard (199)}} & \multicolumn{2}{c|}{\textbf{Solid (200)}} & \multicolumn{2}{c}{\textbf{Average}}\\
			\cline{2-13}
			& \textbf{PSNR} & \textbf{SSIM} & \textbf{PSNR} & \textbf{SSIM} & \textbf{PSNR} & \textbf{SSIM} & \textbf{PSNR} & \textbf{SSIM} & \textbf{PSNR} & \textbf{SSIM} & \textbf{PSNR} & \textbf{SSIM}   \\
			\hline
			BDN (ECCV'18) & 18.83 & 0.737 & 18.68 & 0.728 & 22.02 & 0.822 & 20.54 & 0.857 & 22.68 & 0.856 & 20.55 & 0.800 \\
			ERRNet (CVPR'19) & 20.39 & 0.766 & 22.28 & 0.796 & 25.14 & 0.873 & 21.53 & 0.877 & 23.55 & 0.880 & 22.58 & 0.838  \\
			IBCLN (CVPR'20) & 23.78 & 0.784 & 21.59 & 0.764 & 24.46 & 0.885 & 22.95 & 0.875 & \underline{24.74} & 0.893 & 23.50 & 0.840  \\
			LANet (ICCV'21) & 23.55 & 0.811 & 22.51 & \textbf{0.815} & \underline{26.06} & 0.900 & \textbf{24.14} & \textbf{0.907} & 24.30 & \underline{0.898} & 24.11 & \underline{0.866} \\
			YTMT (NIPS'21) & 24.08 & 0.814 & 22.68 & 0.798 & 25.24 & 0.888 & 21.86 & 0.880 & 23.79 & 0.887 & 23.53 & 0.853  \\
			DMGN (TIP'21) & 20.63 & 0.764 & 20.28 & 0.763 & 21.34 & 0.774 & 22.65 & 0.879 & 23.27 & 0.872 & 21.63 & 0.810  \\
			DSRNet  (ICCV'23) & 24.84 & 0.823 & 22.09 & 0.790 & 26.00 & \underline{0.902} & 20.05 & 0.883 & 23.96 & 0.887 & 23.39 & 0.857  \\
			RDRNet (CVPR'24) & \underline{25.33} & \textbf{0.835} & \underline{22.76} & 0.804 & \textbf{26.97} & \textbf{0.905} & 22.11 & 0.881 & 24.42 & 0.892 & \underline{24.32} & 0.863 \\
			ALANet (Ours) & \textbf{25.56} & \underline{0.829} & \textbf{23.89} & \underline{0.812} & 25.93 & 0.900 & \underline{23.47} & \underline{0.897} & \textbf{24.85} & \textbf{0.900} & \textbf{24.74} & \textbf{0.868} \\
			\hline
		\end{tabular}
	}
	\caption{Quantitative comparison with SOTA methods on public datasets. Bold and underline indicate top 1st and 2nd rank, respectively.}
	\label{tab:quantitative_comparison}
\end{table*}

\begin{figure*}[t]
	\centering
	\includegraphics[width=0.85\textwidth]{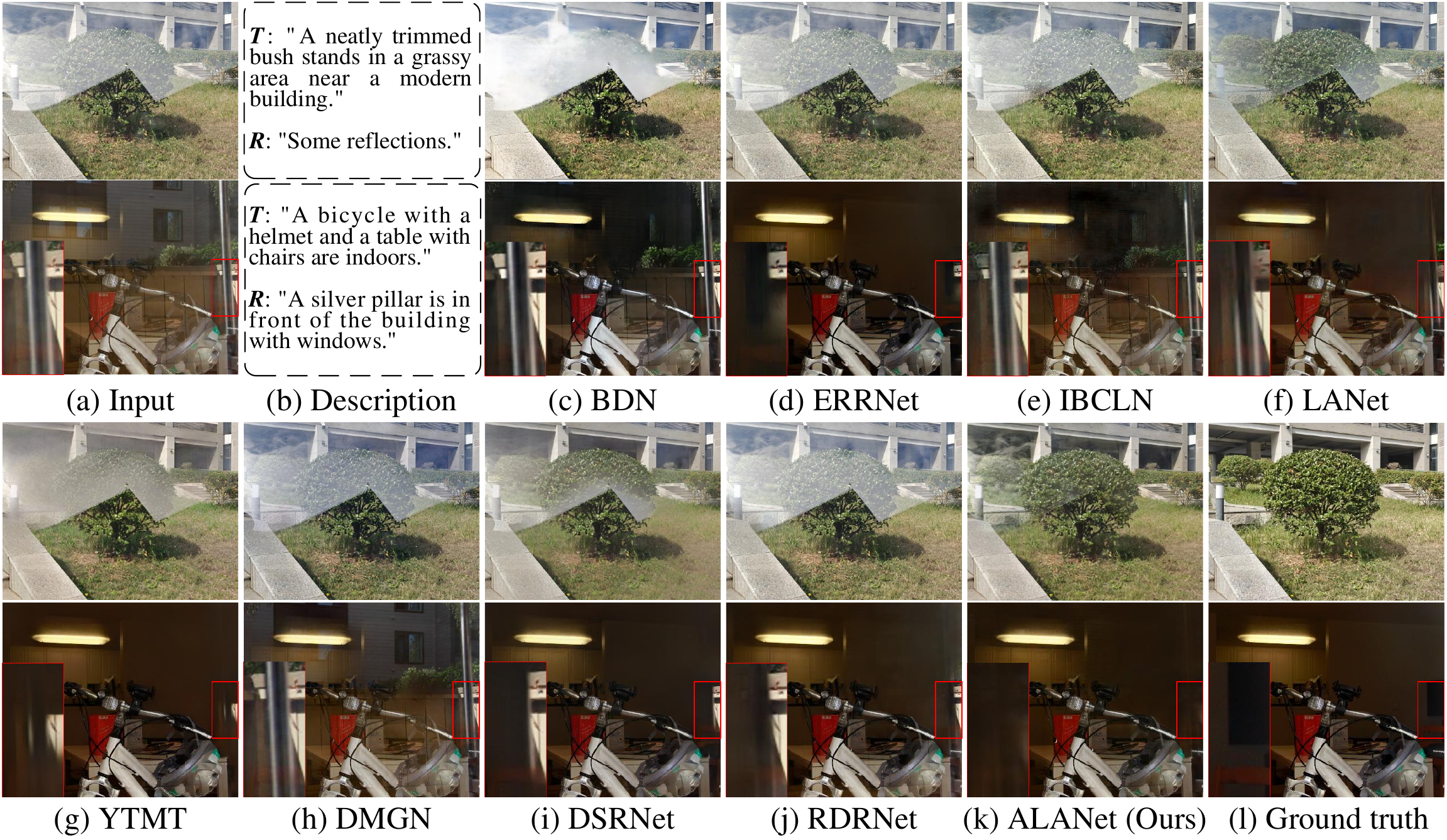} 
	\caption{Qualitative comparison with SOTA methods on the CRLAV dataset (top) and the Real dataset (bottom). ALANet excels in identifying and removing reflections in complex environments. \textbf{\textit{T}} and \textbf{\textit{R}} represent the transmission and reflection layers, respectively.}
	\label{fig7}
\end{figure*}
The Language-Guided Spatial-Channel Cross Transformer (LSCT), as shown in Figure \ref{fig6}, features the Language-Guided Spatial-Channel Cross Attention (LSCA) as its core. The LSCA utilizes the semantic information from language to interact with the spatial and channel dimensions of the feature map, leveraging language-driven attention mechanisms to decouple specific information. As illustrated, the LSCA first applies spatial and channel pooling to the input image features $\bm{F}_I$, obtaining $\bm{F}_S \in \mathbb{R}^{1 \times C}$ and $\bm{F}_C \in \mathbb{R}^{HW \times 1}$, where $HW = H \times W$. The language features $\bm{F}_L$ interact with $\bm{F}_S$ to generate $\bm{M}_{SL} \in \mathbb{R}^{C \times C}$. Each element $(i,j)$ in $\bm{M}_{SL}$ represents the correlation between the $i$-th channel of the image features and the $j$-th channel of the language features, indicating a global language-image interaction effect. Simultaneously, the language features $\bm{F}_L$ interact with $\bm{F}_C$ to produce $\bm{M}_{LC} \in \mathbb{R}^{HW \times C}$. Each element $(h,c)$ in $\bm{M}_{LC}$  indicates the correlation between the $h$-th spatial position of the image and the $c$-th channel of the language features, reflecting the relevance of the language description to specific local regions in the image. Finally, both $\bm{M}_{LC}$ and $\bm{M}_{SL}$ undergo softmax and are multiplied to obtain $\bm{M}_{LCSL} \in \mathbb{R}^{HW \times C}$. This matrix combines global language guidance with local image features, enhancing the model's ability to interpret language descriptions, and focus on specific regions of the image that correspond to these descriptions.

\section{Complex Reflection and Language Accuracy Variance Dataset}
To evaluate the model's performance under complex reflections and varying levels of language accuracy, we propose the Complex Reflection and Language Accuracy Variance (CRLAV) dataset. The dataset encompasses real-world indoor and outdoor scenes, comprising a total of 600 image pairs. Using a tripod-mounted smartphone, we capture images with reflection artifacts by placing glass and acrylic sheets of varying thicknesses as obstructions. Ground truth images without reflections are obtained by removing these obstructions. To achieve the three key characteristics of complex reflections—high intensity, large coverage, and indistinguishability—we employ several strategies during data collection. By adjusting the tilt angles of the obstructions, we control the area and intensity of the reflections, allowing them to cover larger regions and increase the overlap between the reflected light and the transmission layer content, thereby creating high-intensity and large-area reflection artifacts. Additionally, we select complex scenes with rich textures or multiple objects to enhance the confusion and complexity of the reflected content, making it more challenging to distinguish from the actual transmission content. 

To assess model performance under language conditions of varying accuracy, each image is annotated with both accurate and inaccurate language descriptions. Inaccurate descriptions are categorized into three types: incorrect, confused, and incomplete. Each type is further divided into four levels—slightly, moderately, severely, and entirely inaccurate—corresponding to adjustments of 25\%, 50\%, 75\%, and 100\% of the label content, respectively. This setup simulates the impact of language errors on reflection removal models. 
\section{Experiments}
\subsection{Implementation Details}
To balance performance and parameter count, the channel numbers from level 0 to level 4 of the network are set to $C_0, C_1, C_2, C_3, C_4 = \left[ 64, 128, 128, 160, 160 \right]$, and the number of LASBs at each level ($N_0$ to $N_4$) is set to 2. The initial learning rate is $10^{-4}$, with a batch size of 1, a patch size of 224$\times$224, and random flipping applied for data augmentation. The model is trained for 70 epochs using the Adam optimizer \cite{kingma2014adam} with a single RTX 3080 Ti GPU. The learning rate decreases to $10^{-5}$ at 50 epochs.
\subsection{Dataset and Evaluation Metrics}
We employ both synthetic and real-world images to train our model. For synthetic images, we generate data using the popular image captioning dataset Flickr8k \cite{hodosh2013framing}, which contains 8,091 images, each with five different language descriptions. We randomly select images from the Flickr8k dataset to serve as the transmission and reflection layers. These are combined through linear blending \cite{zhang2018single} to generate synthetic images. For the real-world training data, following prior works \cite{zhong2024language,hu2023single,zhu2023hue}, we train our model using 200 image pairs from the Nature dataset \cite{li2020single} and 90 image pairs from the Real dataset \cite{zhang2018single}. 

We use the remaining images from the Nature and Real datasets, along with the three subsets—Wild,  Postcard, and Solid—from the SIR\textsuperscript{2}
dataset \cite{wan2017benchmarking} as public test sets. The CRLAV dataset is also included as a test set. Following prior works \cite{wei2019single,hu2021trash}, to prevent  memory overload, we resize the images in the Real test set by setting the longer side length to 420 while preserving the original aspect ratio. We use the commonly employed peak signal-to-noise ratio (PSNR) \cite{huynh2008scope} and structural similarity index measure (SSIM) \cite{wang2004image} as evaluation metrics, which are calculated in the RGB color space. Higher values indicate better performance.
\begin{table}[t]
	\centering
	\setlength{\tabcolsep}{10pt} 
	\resizebox{\columnwidth}{!}{
		\begin{tabular}{c|cc|cc}
			\hline
			\multirow{2}{*}{\textbf{Methods}} & \multicolumn{2}{c|}{\textbf{CRLAV (600)}} & \textbf{Param} & \textbf{FLOPs} \\
			\cline{2-3}
			& \textbf{PSNR} & \textbf{SSIM} &\textbf{(M)} & \textbf{(G)} \\
			\hline
			BDN (ECCV'18)   & 17.46 & 0.686 & 75.16 & 12.70 \\
			ERRNet (CVPR'19) & 18.93 & 0.702 & 18.95 & 116.72 \\
			IBCLN (CVPR'20) & 18.81 & 0.701 & 21.61 & 98.16 \\
			LANet (ICCV'21) & 19.28 & \underline{0.709} & 10.93 & 83.81 \\
			YTMT (NIPS'21)  & 18.92 & 0.697 & 76.90 & 110.98 \\
			DMGN (TIP'21)   & 18.49 & 0.698 & 45.49 & 116.85 \\
			DSRNet (ICCV'23) & 18.58 & 0.693 & 9.87  & 32.34 \\
			RDRNet (CVPR'24) & \underline{19.51} & 0.706 & 29.09 & 5.14 \\
			\textbf{ALANet (Ours)} & \textbf{19.68} & \textbf{0.719} & 8.69  & 32.92 \\
			\hline
		\end{tabular}
	}
	\caption{Quantitative comparison with SOTA methods on the CRLAV dataset, including parameters and FLOPs (computed for a 128×128 RGB image).}
	\label{tab:param}
\end{table}
\subsection{Comparison Results}
To assess the performance of the proposed ALANet, we compared it with eight SOTA methods: BDN \cite{yang2018seeing}, ERRNet \cite{wei2019single}, IBCLN \cite{li2020single}, LANet \cite{dong2021location}, YTMT \cite{hu2021trash}, DMGN \cite{feng2021deep}, DSRNet \cite{hu2023single}, and RDRNet \cite{zhu2024revisiting}. To ensure a fair comparison, for methods with publicly available training code, we fine-tuned their models on our training datasets and selected the version that performed best. Quantitative comparison results across public datasets are shown in Table \ref{tab:quantitative_comparison}. It can be observed that our ALANet achieves the best or second-best results across multiple datasets, ultimately delivering the best average performance. This demonstrates ALANet's advantage in image reflection removal, confirming its effectiveness and reliability in various reflection scenarios. 


To compare the capability of removing complex reflections, Table \ref{tab:param} presents a quantitative comparison between ALANet and other methods on the CRLAV dataset, where ALANet achieves the best performance. Table \ref{tab:param} also shows the parameter count and floating point operations (FLOPs), demonstrating that ALANet maintains a relatively balanced parameter count and FLOPs compared to SOTA methods.

\begin{table}[t]
	\centering
	\setlength{\tabcolsep}{2pt} 
	\resizebox{\columnwidth}{!}{
		\begin{tabular}{c c c|c c c|c c}
			\hline
			\multicolumn{3}{c|}{\textbf{Language type for training}} & \multicolumn{3}{c|}{\textbf{Language type for testing}} & \multicolumn{2}{c}{\textbf{Average}} \\
			\hline
			\textbf{Correct} & \textbf{Random} & \textbf{None} & \textbf{Correct} & \textbf{Random} & \textbf{None} & \textbf{PSNR} & \textbf{SSIM} \\
			\hline
			\checkmark &× &× & \checkmark &× &× & \textbf{24.74} & \textbf{0.868} \\
			\checkmark &× &× &× & \checkmark &× & 24.09 & 0.861 \\
			\checkmark &× &× &× &× & \checkmark & 24.02 & 0.856 \\
			×& \checkmark &× & \checkmark &× &× & 24.27 & 0.864 \\
			×& \checkmark &× &× & \checkmark &× & 24.11 & 0.860 \\
			×& \checkmark &× &× &× & \checkmark & 23.83 & 0.852 \\
			×&× & \checkmark &× &× & \checkmark & 24.02 & 0.857 \\
			\hline
		\end{tabular}
	}
	\caption{Ablation experiments for different types of language inputs during training and testing on public datasets.}
	\label{tab:DiffTypes}
\end{table}

\begin{figure}[t]
	\centering
	\includegraphics[width=0.89\columnwidth]{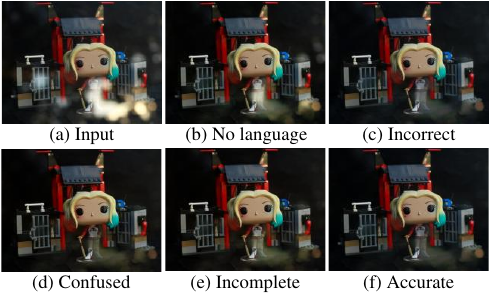} 
	\caption{Visual effects of our ALANet with  different types of language inputs. The specific language inputs for each subfigure are provided in the supplementary material.}
	\label{fig:inaccurate}
\end{figure}
Figure \ref{fig7} presents  a qualitative comparison of ALANet with SOTA methods across various scenes. In the complex outdoor scenario from the CRLAV dataset, only ALANet achieves the most thorough removal of reflections. In the complex indoor scene from the Real dataset, although  ERRNet, DSRNet, and RDRNet can remove reflections from the silver column, they fail to eliminate reflections from the light. Only ALANet successfully removes reflections from both the light and the silver column. This demonstrates that ALANet is more effective at removing complex reflections compared to other methods.

\subsection{Ablation Studies}
In this section, we delve into the impact of different components of ALANet by conducting various ablation studies.

\textbf{Impact of language input}. To explore the impact of language and its accuracy on model performance, we conducted experiments using different types of language inputs during the training and testing phases, with the results shown in Table \ref{tab:DiffTypes}. It can be observed that when correct language is used for training, the performance progressively decreases when tested with correct, random, and no language inputs, indicating that correct language input is most beneficial for enhancing model performance. Simultaneously, even with the lower accuracy caused by random language inputs, benefiting from our filtering and optimization strategies, the performance with random language remains better than with no language input. When training with random language, testing with both correct and random language still achieved commendable performance, surpassing that of RDRNet and DSRNet respectively on the SSIM metric. This demonstrates that random language during training did not mislead the model into confusion, showing that the model can still perform well even under conditions of low language accuracy.

Figure \ref{fig:inaccurate} illustrates the visual  effects of our ALANet in removing complex reflections with three types of inaccurate descriptions: incorrect, confused, and incomplete. It can be seen that ALANet can utilize the portions of inaccurate language inputs that match the semantics of the corresponding layers, resulting in de-reflection outcomes that are superior to those without any language inputs.

Table \ref{tab:language-inaccuracy} illustrates the performance of our ALANet under varying degrees of language accuracy. It can be observed that the
model's performance decreases as the degree of accuracy declines on both public datasets and the CRLAV dataset. Notably, even with severely incorrect, confused, or incomplete input, the model's performance remains superior to that with no language input. This demonstrates ALANet's robustness to severely inaccurate language.

\begin{table}[t]
	\centering
	\setlength{\tabcolsep}{2.5pt} 
	\resizebox{\columnwidth}{!}{
		\begin{tabular}{c|c|cc|cc|cc}
			\hline
			&\multirow{2}{*}{\textbf{Degree}} & \multicolumn{2}{c|}{\textbf{Incorrect}} & \multicolumn{2}{c|}{\textbf{Confused}} & \multicolumn{2}{c}{\textbf{Incomplete}} \\ \cline{3-8}
			&& \textbf{PSNR} & \textbf{SSIM} & \textbf{PSNR} & \textbf{SSIM} & \textbf{PSNR} & \textbf{SSIM} \\ \hline
			\multirow{4}{*}{\makecell{\textbf{Public}\\ \textbf{datasets}\\ \textbf{(494)}}} 
			&Slightly (25\%)   & 24.55 & 0.865 & 24.48 & 0.866 & 24.36 & 0.864 \\
			&Moderately (50\%) & 24.35 & 0.862 & 24.29 & 0.861 & 24.22 & 0.861 \\
			&Severely (75\%)   & 24.08 & 0.859 & 24.01 & 0.859 & 24.03 & 0.857 \\
			&Entirely (100\%)  & 22.70 & 0.838 & 24.01 & 0.854 & 24.02 & 0.856 \\
			\hline
			\multirow{4}{*}{\makecell{\textbf{CRLAV}\\ \textbf{(600)}} } 
			&Slightly (25\%)   & 19.43 & 0.713 & 19.65 & 0.714 & 19.52 & 0.718 \\
			&Moderately (50\%) & 19.25 & 0.707 & 19.59 & 0.712 & 19.33 & 0.716 \\
			&Severely (75\%)   & 19.23 & 0.705 & 19.51 & 0.711 & 19.24 & 0.714 \\
			&Entirely (100\%)  & 19.06 & 0.703 & 19.41 & 0.708 & 19.13 & 0.697 \\
			\hline
		\end{tabular}
	}
	\caption{Ablation experiments on the effects of varying degrees of language inaccuracy.}
	\label{tab:language-inaccuracy}
\end{table}

\begin{table}[t]
	\centering
	\setlength{\tabcolsep}{1pt} 
	\resizebox{\columnwidth}{!}{
		\begin{tabular}{cc|cc|cc|cc}
			\hline
			\multirow{3}{*}{\textbf{ALCM}} & \multirow{3}{*}{\textbf{LSCT}} & \multicolumn{2}{c|}{\textbf{LCAM}} & \multicolumn{2}{c|}{\textbf{Average}} & \multirow{3}{*}{\makecell{\textbf{Param} \\ \textbf{(M)}}} & \multirow{3}{*}{\makecell{\textbf{FLOPs} \\ \textbf{(G)}}} \\ \cline{3-6}
			&  & \textbf{Language-guided} & \textbf{Channel} & \multirow{2}{*}{\textbf{PSNR}} & \multirow{2}{*}{\textbf{SSIM}} &  &  \\ 
			&  & \textbf{attention} & \textbf{attention} & & &  & \\ \hline
			\checkmark & \checkmark & \checkmark & \checkmark & \textbf{24.74} & \textbf{0.868} & 8.69 & 32.92 \\
			\checkmark & \checkmark & \checkmark &×  &  24.52&	0.865&	8.46&	32.91  \\
			\checkmark & \checkmark &×  & \checkmark &  24.05&	0.859&	8.24&	32.92  \\
			\checkmark & \checkmark &×  & × &24.11&0.858&	8.01&32.90 \\
			× &\checkmark  & \checkmark & \checkmark & 24.41 & 0.867 & 8.51 & 32.92 \\
			×& × & \checkmark & \checkmark & 24.19 & 0.863 & 8.20 & 32.89 \\ \hline
		\end{tabular}
	}
	\caption{Ablation experiments for different modules on public datasets. FLOPs for a 128x128 RGB image.}
	\label{tab:modules}
\end{table}


\begin{figure}[t]
	\centering
	\includegraphics[width=0.89\columnwidth]{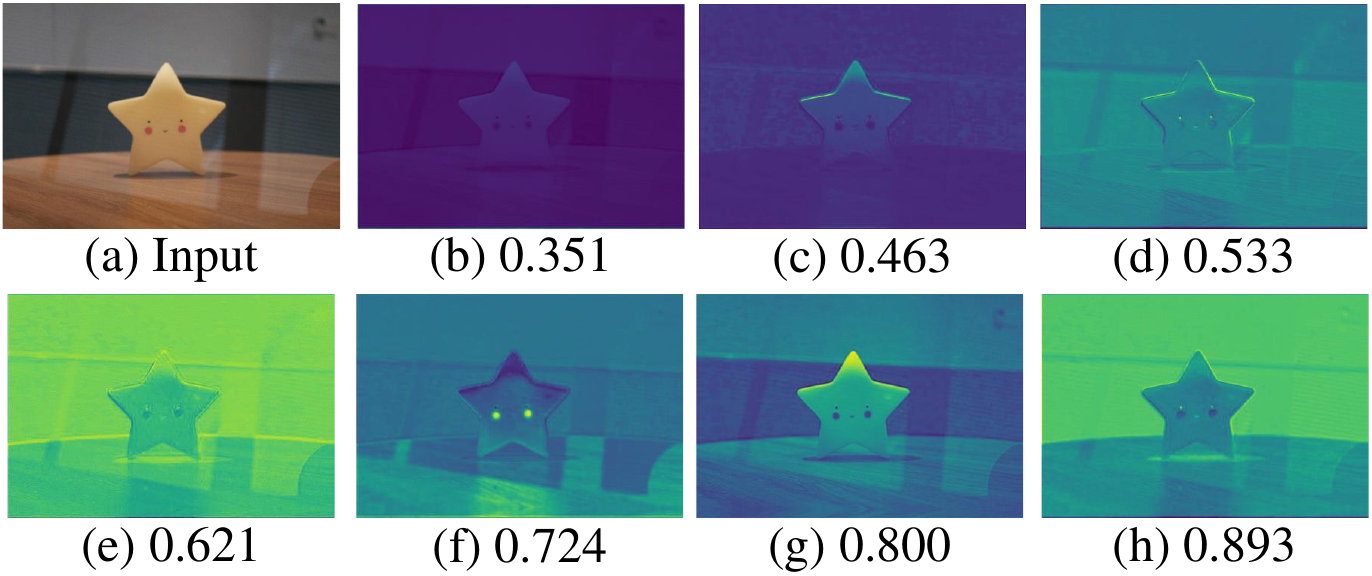} 
	\caption{Feature maps in LCAM and corresponding language-guided attention weights, with the language description: ``A star-shaped toy on the tabletop."}
	\label{fig:channel_weights}
\end{figure}


\begin{figure}[t]
	\centering
	\includegraphics[width=0.89\columnwidth]{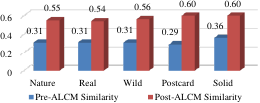} 
	\caption{Comparison of Pre-ALCM and Post-ALCM similarities between language and corresponding image features across different datasets.}
	\label{fig9}
\end{figure}

\textbf{Ablation study on LCAM}. Table \ref{tab:modules} illustrates the contributions of language-guided attention and channel attention to performance within LCAM. It can be observed that the experiments combining both types of attention achieved the best performance, demonstrating the effectiveness of LCAM's competitive attention mechanism. Figure \ref{fig:channel_weights} further shows the feature maps within LCAM and the corresponding language-guided attention weights. It can be observed that the closer the feature map content is to the language description, the higher the language-guided attention weight becomes. This helps to emphasize the regions in the image related to the language description, thereby decoupling the described object.

\textbf{Ablation study on ALCM}. Table \ref{tab:modules} showcases the performance of models with and without ALCM. It is clear that the models equipped with ALCM outperform those without, demonstrating that ALCM enhances the role of language in image reflection removal. To further validate that ALCM can improve the alignment between language and image features, Figure \ref{fig9} displays the cosine similarities between language features and image features before and after processing through ALCM. The similarities post-ALCM are consistently higher across various datasets, indicating an improved alignment between language and image features following the application of ALCM.

\textbf{Ablation study on LSCT}. To demonstrate the effectiveness of LSCT, we conducted experiments by further removing LSCT on top of removing ALCM, with the results shown in Table \ref{tab:modules}. It can be observed that the performance without LSCT is lower than that with LSCT. This indicates that LSCT plays a positive role in enhancing reflection removal performance, especially when paired with ALCM, which amplifies the effects of LSCT.

\section{Conclusion}
In this paper, we propose the ALANet to remove complex reflections with low dependence on language accuracy. Specifically, ALANet employs filtering and optimization strategies to mitigate the effects of inaccurate language and enhance the alignment between language and visual features, while leveraging language cues to decouple layer content from feature maps. Additionally, we introduce the CRLAV dataset to evaluate the model's performance under complex reflections and varying levels of language accuracy. Experimental results demonstrate the effectiveness of the ALANet and its superiority over SOTA methods.

\bibliographystyle{named}
\bibliography{ijcai25}

\clearpage
\appendix
\twocolumn[{%
	\centering
	{\LARGE\bfseries Supplementary Material for Adaptive Language-Aware Image Reflection Removal\par}
	\vspace{0.5em}
	{\LARGE\bfseries Network\par}
	\vspace{1em}
}]
This supplementary material accompanies the paper \textit{Adaptive Language-Aware Image Reflection Removal Network} and provides additional details.

\section{Multi-Receptive Field Decoupling Module}
\begin{figure}[t]
	\centering
	\includegraphics[width=0.98\columnwidth]{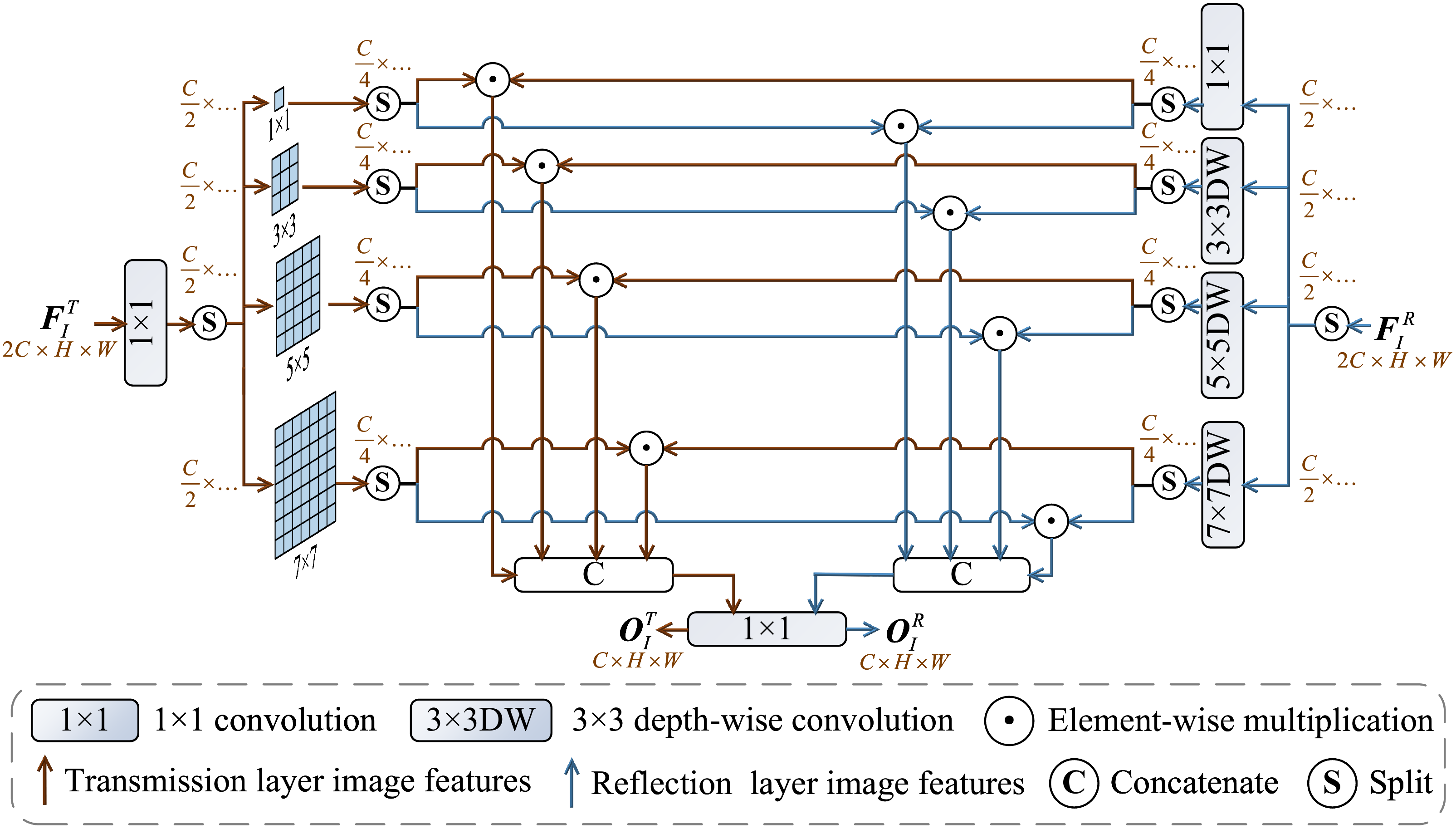} 
	\caption{Multi-receptive field decoupling module.}
	\label{fig1}
\end{figure}

Reflections can appear in different sizes depending on the distance and angle of the reflective surfaces. They can range from large, nearly transparent reflections that cover most of the image to small, highly distorted reflections that only affect a small part of the view. Therefore, we design the Multi-Receptive Field Decoupling Module (MFDM), as shown in Figure \ref{fig1}, to respond to different reflection scales present in the image. The MFDM reduces interference between features of different scales by dividing channels corresponding to each receptive field size. Additionally, the output features of each receptive field are multiplied by the corresponding features of the transmission or reflection layer, achieving targeted feature enhancement and suppressing irrelevant features.

\section{Loss Functions}
To ensure efficient learning, we adopt the Mean Squared Error (MSE) loss $\ell_{\text{MSE}}$, gradient loss $\ell_{\text{grad}}$, and perceptual loss $\ell_{\text{VGG}}$ to train the network. $\ell_{\text{MSE}}$ evaluates the overall pixel-level error between the reconstructed image and the Ground Truth (GT). $\ell_{\text{grad}}$ helps the network identify and restore edge information affected by reflections.  $\ell_{\text{VGG}}$ ensures that the reconstructed image is visually similar to the GT in terms of perceptual quality. By combining these losses, the overall loss function  $\ell_{\text{total}}$ can be expressed as follows:
\begin{equation}
	\begin{aligned}
		\ell_{\text{total}} = & \, \lambda_1 \left( \ell_{\text{MSE}} \left( \bm{\hat{T}}, \bm{T} \right) + \ell_{\text{MSE}} \left( \bm{\hat{R}}, \bm{R} \right) \right) \\
		& + \lambda_2 \ell_{\text{grad}} \left( \bm{\hat{T}}, \bm{T} \right) + \lambda_3 \ell_{\text{VGG}} \left( \bm{\hat{T}}, \bm{T} \right)
	\end{aligned}
\end{equation}
where $\bm{\hat{T}}$ and $\bm{\hat{R}}$ are the network outputs for the reconstructed transmission layer and reflection layer, respectively, and $\bm{T}$ and $\bm{R}$ are their corresponding GTs. The parameters $\lambda_1$, $\lambda_2$, and $\lambda_3$ are weight parameters used to adjust the contribution ratio of each loss term. 

\section{Supplemental Ablation Studies}
\subsubsection{Analysis of Language-Aware Competition Attention Module (LCAM).}
To analyze how LCAM leverages language information to enhance the performance of reflection removal, Figure \ref{fig2} visualizes the effects of different attentions during the separation of the transmission layer by LCAM. It can be observed that the channel attention shown in (b) primarily focuses on all content within the image, with limited capability to distinguish between transmission and reflection contents. The language-guided attention map displayed in (c) divides the image content into building and sky parts by the linguistic cues and focuses on the former. Channel attention and language-guided attention competitively merge, resulting in the language-aware competition attention map in (d), which primarily displays the content of the transmission layer and effectively minimizes reflection content.
\begin{figure}[t]
	\centering
	\includegraphics[width=0.9\columnwidth]{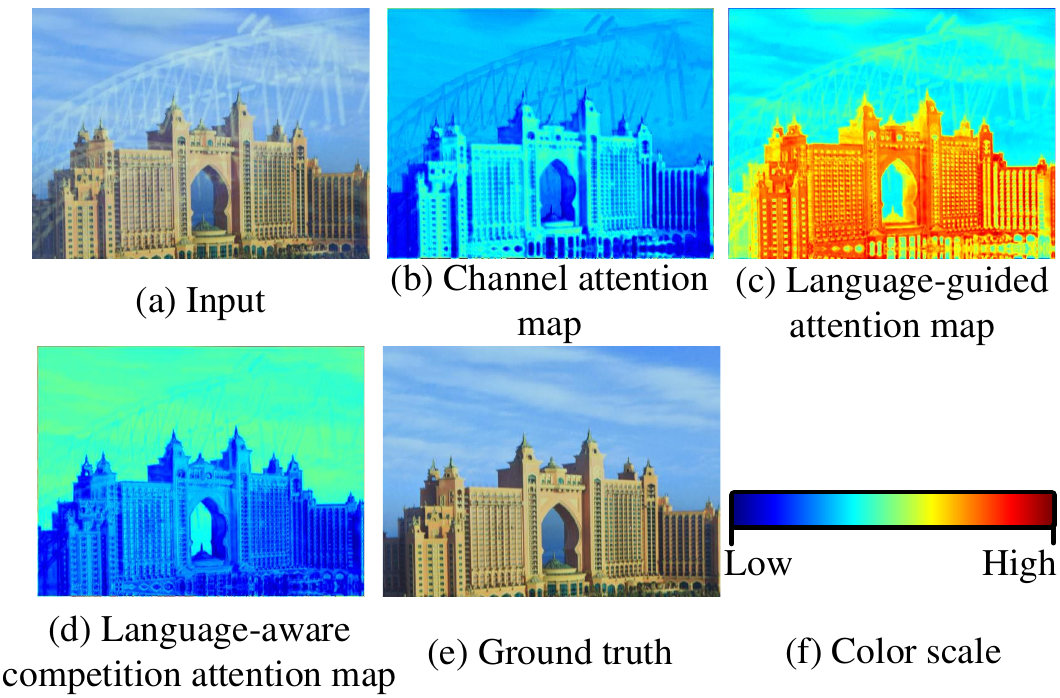} %
	\caption{Internal attention maps of the LCAM for transmission layer separation. Language description: ``A building under the sky."}
	\label{fig2}
\end{figure}
\subsubsection{Ablation study on the number of language inputs.} 
\begin{figure}[t]
	\centering
	\includegraphics[width=0.9\columnwidth]{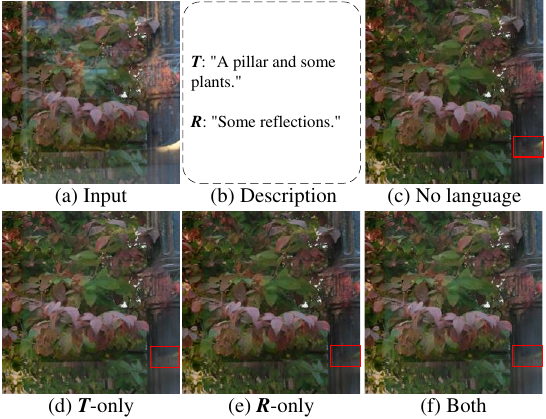} 
	\caption{Effects of our ALANet with different language numbers. \textbf{\textit{T}} and \textbf{\textit{R}} represent the transmission and reflection layers, respectively.}
	\label{fig_langType}
\end{figure}

Table \ref{tab:language-inputs} and Figure \ref{fig_langType} demonstrate the impact of different numbers of language inputs on the model's performance. The results show that the model performs best when both transmission layer and reflection layer language inputs are provided, indicating that comprehensive language input is most beneficial for the model to distinguish between different layer contents. When only transmission layer or reflection layer language input is provided, the performance is not as high as with both inputs, but it still surpasses the performance with no language input. This suggests that even partial language guidance can enhance the model's ability to differentiate between layers, highlighting the effectiveness of language in improving the model's performance.



\begin{table}[t]
	\centering
	\setlength{\tabcolsep}{7pt} 
	\resizebox{\columnwidth}{!}{
		\begin{tabular}{c c|c c}
			\hline
			\multicolumn{2}{c|}{\textbf{Language input}} & \multicolumn{2}{c}{\textbf{Average}} \\ \cline{1-4}
			\textbf{Transmission layer} & \textbf{Reflection layer} & \textbf{PSNR} & \textbf{SSIM} \\ \hline
			\checkmark & \checkmark  & \textbf{24.74} & \textbf{0.868} \\ 
			\checkmark & ×  & 24.60 & 0.866 \\ 
			×& \checkmark & 24.36 & 0.862 \\ 
			×& × & 24.02 & 0.856 \\ \hline
		\end{tabular}
	}
	\caption{Ablation experiments on the number of language inputs.}
	\label{tab:language-inputs}
\end{table}

\subsubsection{Ablation study on channel numbers.}
Table \ref{tab:channel-numbers} shows the model's performance with different channel numbers. It demonstrates that setting the number of channels to 64, 128, 128, 160, and 160 for levels 0 through 4 achieves the optimal balance between performance and cost.

\begin{table}[t]
	\centering
	
	\setlength{\tabcolsep}{2pt} 
	\resizebox{\columnwidth}{!}{
		\begin{tabular}{ccccc|cc|cc}
			\hline
			\textbf{Level} & \textbf{Level} & \textbf{Level} & \textbf{Level} & \textbf{Level} &  \multicolumn{2}{c|}{\textbf{Average}}&
			\textbf{Param} &\textbf{FLOPs} \\ \cline{6-7}
			\textbf{0} & \textbf{1} & \textbf{2} & \textbf{3} & \textbf{4} &
			\textbf{PSNR} & \textbf{SSIM}&\textbf{(M)} & \textbf{(G)}\\
			\hline
			64 & 128 & 128 & 160 & 160 & \textbf{24.74} & \textbf{0.868}& 8.69&32.92 \\
			64 & 128 & 256 & 512 & 512 & 24.53 & 0.866&55.71&53.36 \\
			64 & 160 & 160 & 160 & 160 & 24.35 &	0.866&10.36&40.32 \\
			64 & 128 & 160 & 160 & 160 & 24.38 & 0.867&9.54&34.56 \\
			64 & 128 & 128 & 128 & 128 & 24.27 & 0.865&6.95&32.42 \\
			\hline
		\end{tabular}
	}
	\caption{Ablation experiments on channel numbers.}
	\label{tab:channel-numbers}
\end{table}

\subsubsection{Ablation study on loss functions.}
Table \ref{tab:loss-functions} presents the model's performance with different loss weights: MSE loss weight $\lambda_1$, gradient loss weight $\lambda_2$, and perceptual loss weight $\lambda_3$. It shows that setting $\lambda_1$, $\lambda_2$, and $\lambda_3$ to 1.0, 2.0, and 0.01, respectively, achieves optimal performance.

\begin{table}[t]
	\centering
	
	\setlength{\tabcolsep}{5pt} 
	\resizebox{\columnwidth}{!}{
		\begin{tabular}{cccccc}
			\hline
			\textbf{MSE loss}  & \textbf{gradient loss}  & \textbf{perceptual loss} & \multicolumn{2}{c}{\textbf{Average}}  \\ 
			\cline{4-5}
			\bm{$\lambda_1$} &\bm{$\lambda_2$} &\bm{$\lambda_3$}& \textbf{PSNR} & \textbf{SSIM}  \\
			\hline
			1.0 & 2.0 & 0.01 & \textbf{24.74} & \textbf{0.868} & \\
			1.0 & 1.0 & 0.01 & 24.57 & 0.867 & \\
			1.0 & 2.0 & 0.02 & 24.49 & 0.867 & \\
			1.0 & 2.0 & 0.0 & 24.36	 & 0.864 \\
			1.0 & 0.0 & 0.0  & 24.11 & 0.854 & \\
			\hline
		\end{tabular}
	}
	\caption{Ablation experiments on loss functions.}
	\label{tab:loss-functions}
\end{table}

\section{Further Implementation Details}
The model is trained using an Intel(R) Core(TM) i7-6800K CPU @ 3.40GHz, running on Ubuntu 18.04.5 LTS, with Python 3.7.1 and PyTorch 1.9.1. During the training process, we mix the Flickr8k, Nature-200, and Real-90 datasets in a ratio of [0.6, 0.2, 0.2] for hybrid training.

{Additional explanations for Figure 1 in the main paper}.
The incorrect inputs are ``A man riding a skateboard down a sidewalk'', which is generated by BLIP \cite{li2022blip} from the input image for \textbf{\textit{T}}, and ``Some reflections'', which is a manually provided description for \textbf{\textit{R}}.  
The confused inputs are ``A road and some windows'' for \textbf{\textit{T}}, and ``Some trees and a building'' for \textbf{\textit{R}}.  
The incomplete inputs are ``A road'' for \textbf{\textit{T}}, and ``Windows'' for \textbf{\textit{R}}.  
The accurate inputs are ``Some trees beside a road and a building'' for \textbf{\textit{T}}, and ``A building with windows'' for \textbf{\textit{R}}.  

{Additional explanations for Figure 8 in the main paper}.
The incorrect inputs are ``A toy made of a car and a gas station'' for \textbf{\textit{T}}, and ``A string of bells'' for \textbf{\textit{R}}.  
The confused inputs are ``Building blocks'' for \textbf{\textit{T}}, and ``A toy made of a doll and light'' for \textbf{\textit{R}}.  
The incomplete inputs are ``Building blocks'' for \textbf{\textit{T}}, and ``Light'' for \textbf{\textit{R}}.  
The accurate inputs are ``A toy made of a doll and building blocks'' for \textbf{\textit{T}}, and ``Some light spots'' for \textbf{\textit{R}}.



\section{Comparison with Language-guided Reflection Removal Methods.}
\begin{figure*}[t]
	\centering
	\includegraphics[width=0.9\textwidth]{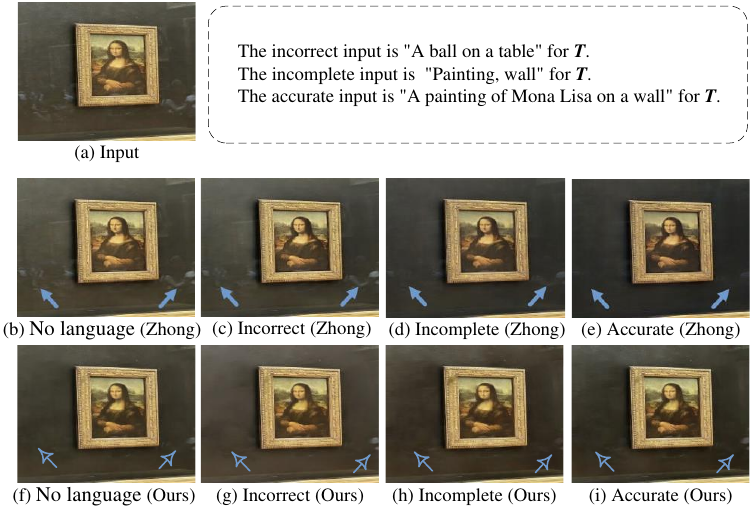} 
	\caption{Qualitative comparison with Zhong et al.'s method using different types of language inputs.}
	\label{fig_compareZhong1}
\end{figure*}

Zhong et al.'s method \cite{zhong2024language} and L-DiffER \cite{hong2025differ} are currently the methods that employ language guidance for image reflection removal. Since their code and introduced datasets are not yet publicly available, we conducted comparisons using relevant examples from Zhong et al.'s paper, as shown in Figure \ref{fig_compareZhong1}. It can be seen that Zhong et al.'s method fails noticeably in reflection removal with no language input or incorrect language input. In contrast, our ALANet  outperforms Zhong et al.'s in cases of no language input, incorrect language input, and incomplete language input. This highlights the advantage of our ALANet in effectively removing reflections even without language guidance or with inaccurate language guidance.

\section{Additional Comparisons}
In this section, we provide additional visual comparisons, as shown in Figures \ref{figv1}-\ref{figv4}, to further demonstrate the effectiveness of our ALANet in removing reflections from images. These comparisons are made against State-of-the-Art (SOTA) methods, including BDN \cite{yang2018seeing}, ERRNet \cite{wei2019single}, IBCLN \cite{li2020single}, LANet \cite{dong2021location}, YTMT \cite{hu2021trash}, DMGN \cite{feng2021deep}, DSRNet \cite{hu2023single}, and RDRNet \cite{zhu2024revisiting}. By evaluating a variety of challenging scenarios, we highlight ALANet's superior ability to remove reflections while preserving image details more effectively than competing methods.

To assess alignment with human visual preferences, we invite 40 participants to evaluate results from ALANet and the compared methods on 10 images (2 from Figure~7 in the main paper and 8 from Figures~6–11 in this supplementary material), with presentation order randomized. Each participant selects their Top-1, Top-2, and Top-3 choices. ALANet receives 16 Top-1 votes, the highest among all methods, exceeding RDRNet’s 12 votes in second place. Overall, ALANet obtains 39 votes across the Top-1, Top-2, and Top-3 rankings, surpassing RDRNet’s 30 votes. These results indicate that ALANet’s outputs better align with human visual preferences.

To demonstrate the robustness of ALANet to inaccurate language descriptions, Figures \ref{compare2_varyinglanguageaccuracies}-\ref{compare1_varyinglanguageaccuracies} illustrate ALANet's performance with various inaccurate language conditions. It can be observed that even with moderately or severely inaccurate descriptions, ALANet still removes reflections more effectively than SOTA methods.

\section{CRLAV Dataset}

The CRLAV dataset contains real-world complex reflections from a variety of scenes, challenging the models' ability to remove these complex reflections. Figure \ref{figv6} shows image examples from the CRLAV dataset.

The CRLAV dataset contains language labels with varying levels of inaccuracy. To simulate types of inaccurate language, a set proportion of nouns are replaced with "error," verbs with "misdo," adjectives with "incorrect," adverbs with "incorrectly," prepositions with random prepositions, and numerals with random numbers in the language labels. To simulate confused language types, words in the language labels for the transmission and reflection layers are randomly exchanged at a set proportion. To simulate incomplete language types, a set proportion of words in the language labels are randomly removed. Additionally, entirely incomplete language means no language input, entirely confused language means completely exchanging the language labels between the transmission and reflection layers, and entirely incorrect language means replacing the entire language label with "incorrect sentence." Figure \ref{dataset_butongzhunquedu} demonstrates image examples from the CRLAV dataset along with corresponding language labels of varying accuracy.
\begin{figure}[t]
	\centering
	\includegraphics[width=0.98\columnwidth]{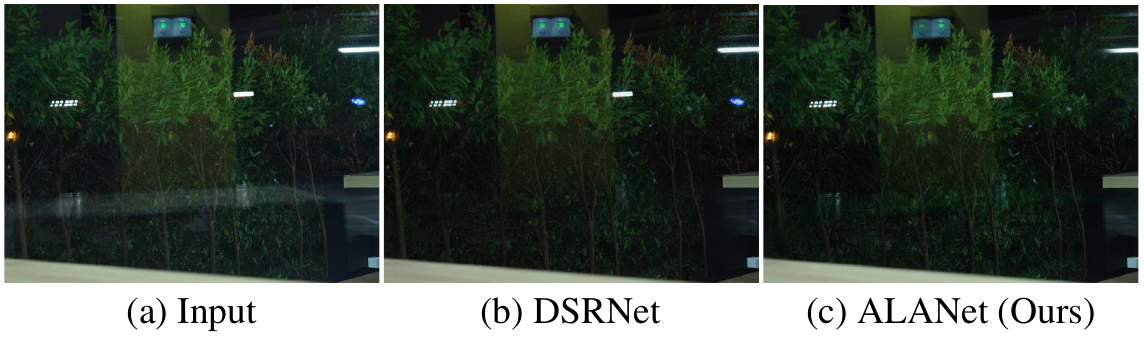} 
	\caption{Failure case of strong nighttime indoor reflection.}
	\label{fig10}
\end{figure}

The accuracy of these language labels is quantitatively evaluated using ROUGE-1 (R-1), ROUGE-L (R-L)~\cite{lin2004rouge}, and METEOR (ME)~\cite{banerjee2005meteor}, which are widely adopted in natural language processing. As reported in Table~\ref{tab:Quantitative Assessment of Language Descriptions}, the measured scores are consistent with the accuracy levels of the dataset labels.

Table~\ref{tab:dataset-comparison} compares the CRLAV dataset with existing publicly available reflection removal datasets, including SIR\textsuperscript{2}~\cite{wan2017benchmarking}, Real~\cite{zhang2018single}, Nature~\cite{li2020single}, and RRW~\cite{zhu2024revisiting}. Most existing datasets do not specifically target complex reflections, whereas CRLAV consists entirely of complex reflection scenes, allowing for more targeted and challenging evaluation. Additionally, publicly available datasets lack language labels, and do not account for the impact of language accuracy on model performance. CRLAV addresses these gaps by including language labels with varying accuracy levels, enabling the assessment of the model’s robustness under guidance with different language accuracies.

To provide richer language evaluation, we further propose an extended version of CRLAV, called CRLAV-plus, where each image’s transmission layer is annotated with five captions, and the reflection layer with one or multiple captions.

\begin{table}[t]
	\centering
	
	\setlength{\tabcolsep}{2.5pt} 
	\resizebox{\columnwidth}{!}{
		\begin{tabular}{c|c c c|ccc|ccc}
			\hline
			\multirow{2}{*}{\textbf{CRLAV}} & \multicolumn{3}{c|}{\textbf{Incorrect}} & \multicolumn{3}{c|}{\textbf{Confused}} & \multicolumn{3}{c}{\textbf{Incomplete}} \\ \cline{2-10}
			& R-1 & R-L & ME & R-1 & R-L & ME & R-1 & R-L & ME \\ \hline
			Slightly & 0.854 & 0.854 & 0.849 & 0.854 & 0.853 & 0.848 & 0.915 & 0.915 & 0.858 \\ 
			Moderately & 0.494 & 0.493 & 0.479 & 0.498 & 0.495 & 0.475 & 0.659 & 0.659 & 0.495 \\ 
			Severely & 0.153 & 0.151 & 0.207 & 0.156 & 0.153 & 0.209 & 0.233 & 0.233 & 0.220 \\ \hline
		\end{tabular}
	}
	\caption{Quantitative assessment of language accuracy. Higher metric values indicate greater accuracy}
	\label{tab:Quantitative Assessment of Language Descriptions}
\end{table}

\begin{table}[t]
	\centering
	\setlength{\tabcolsep}{2pt} 
	\resizebox{\columnwidth}{!}{
		\begin{tabular}{c|cccc|cc}
			\hline
			\textbf{Dataset} & \textbf{Year} & \textbf{Pairs} & \textbf{Scenes} & \textbf{Reflection Types} & \textbf{Language} & \textbf{Accuracy} \\ \hline
			SIR$^2$  & 2017 & 454 & 115 & Not Specified & × & - \\ 
			Real  & 2018 & 109 & 109 & Not Specified & × & - \\ 
			Nature & 2020 & 220 & 68 & Not Specified & × & - \\ 
			RRW   & 2024 & 14952 & 150 & Not Specified & × & - \\ 
			CRLAV (Ours) & 2025 & 600 & 163 & Complex & \checkmark & Varying \\ \hline
		\end{tabular}
	}
	\caption{Comparison of real-world reflection removal datasets.}
	\label{tab:dataset-comparison}
\end{table}

\section{Limitations}
Our ALANet faces challenges in handling some strong indoor reflections at night, as shown in Figure \ref{fig10}. This is because when the outdoor light transmitted at night is weaker than the indoor lighting, the glass windows reflect more of the indoor lighting. Consequently, the outdoor scenery is visually obscured by the indoor reflections, making it difficult to separate the transmission and reflection layers. In the future, we will consider the differences in indoor and outdoor lighting conditions at night, enabling the model to adapt to various lighting conditions.

\begin{figure*}[t]
	\centering
	\includegraphics[width=0.9\textwidth]{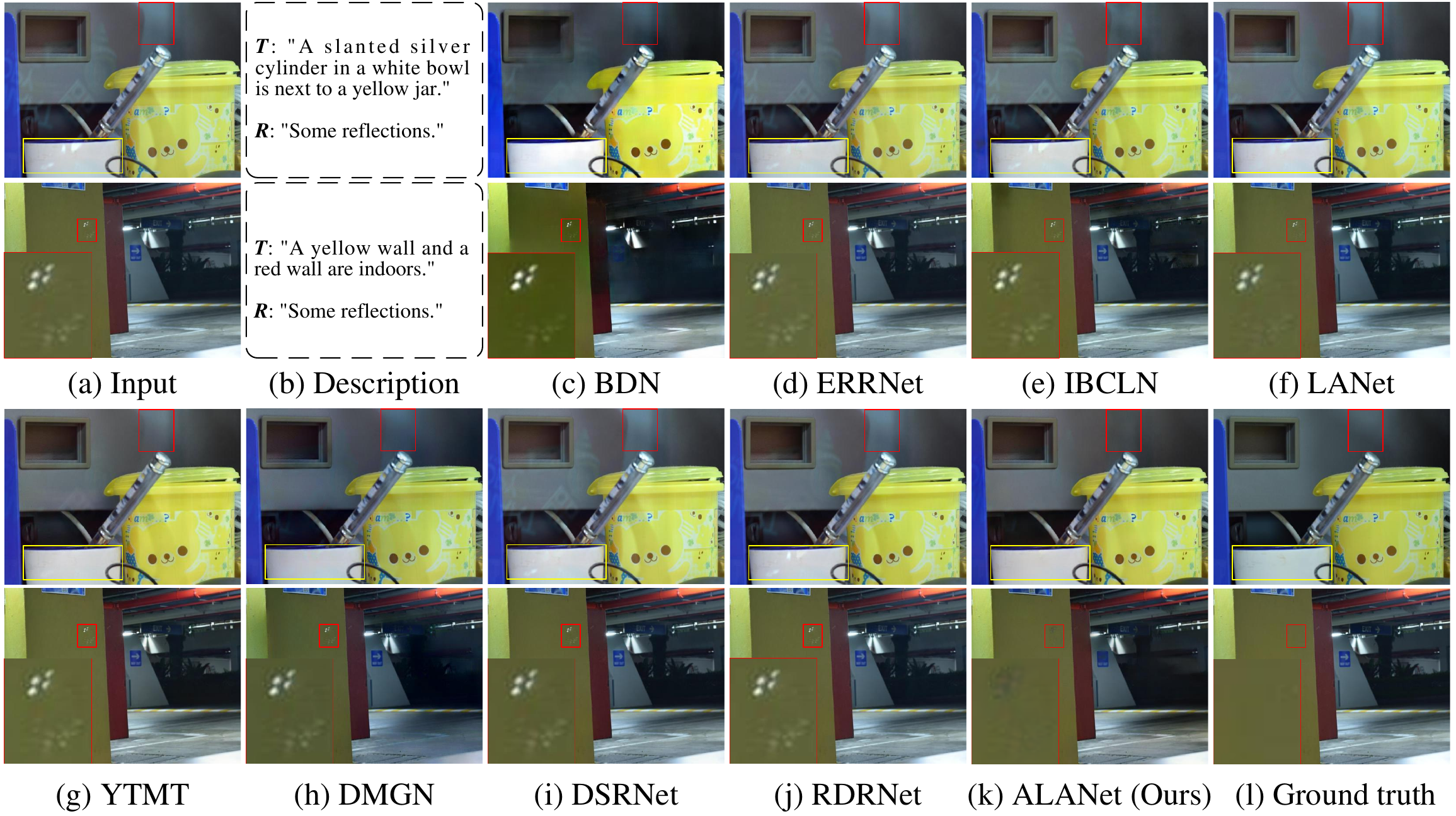} 
	\caption{Qualitative comparison with SOTA methods on the Solid dataset (top) and the Wild dataset(bottom).}
	\label{figv1}
\end{figure*}

\begin{figure*}[t]
	\centering
	\includegraphics[width=0.9\textwidth]{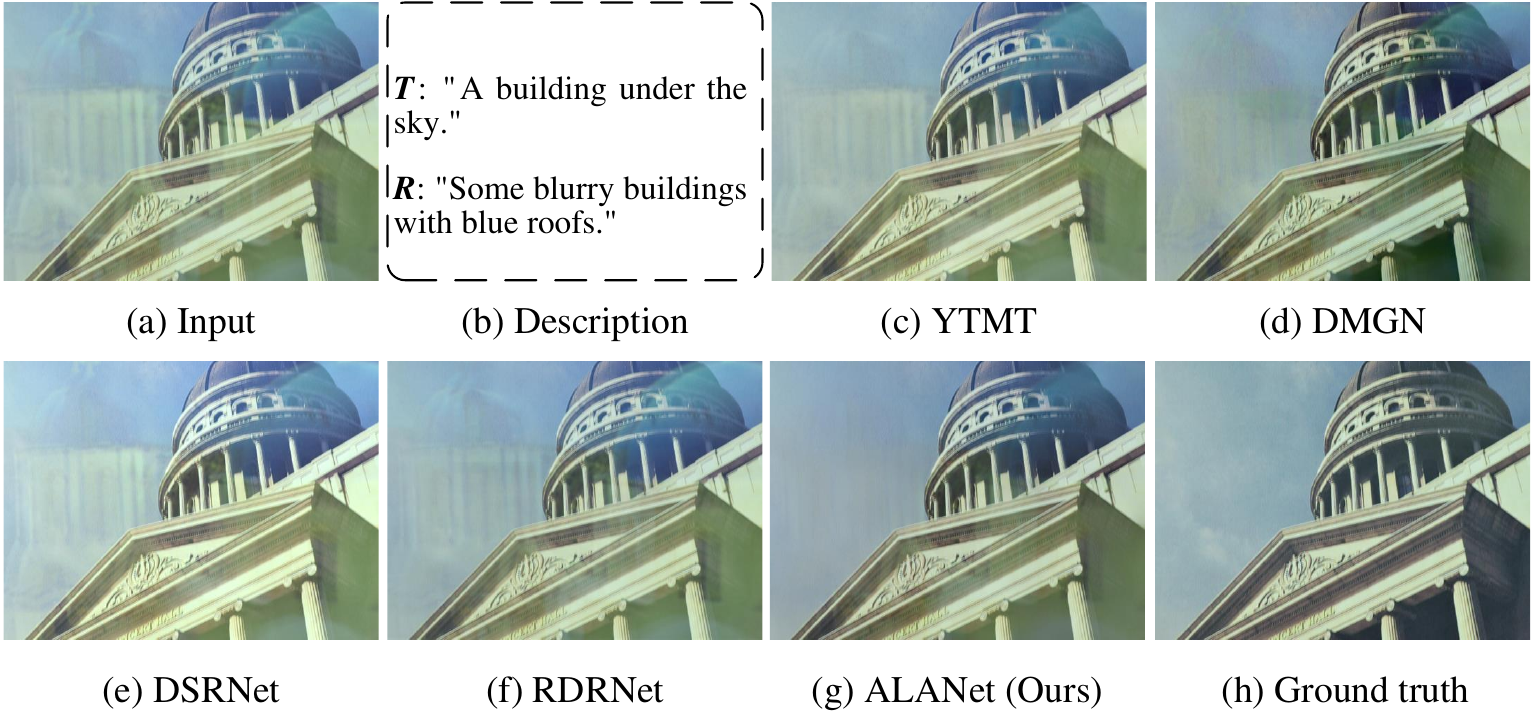} 
	\caption{Qualitative comparison with SOTA methods on the Postcard dataset.}
	\label{figv2}
\end{figure*}

\begin{figure*}[t]
	\centering
	\includegraphics[width=0.9\textwidth]{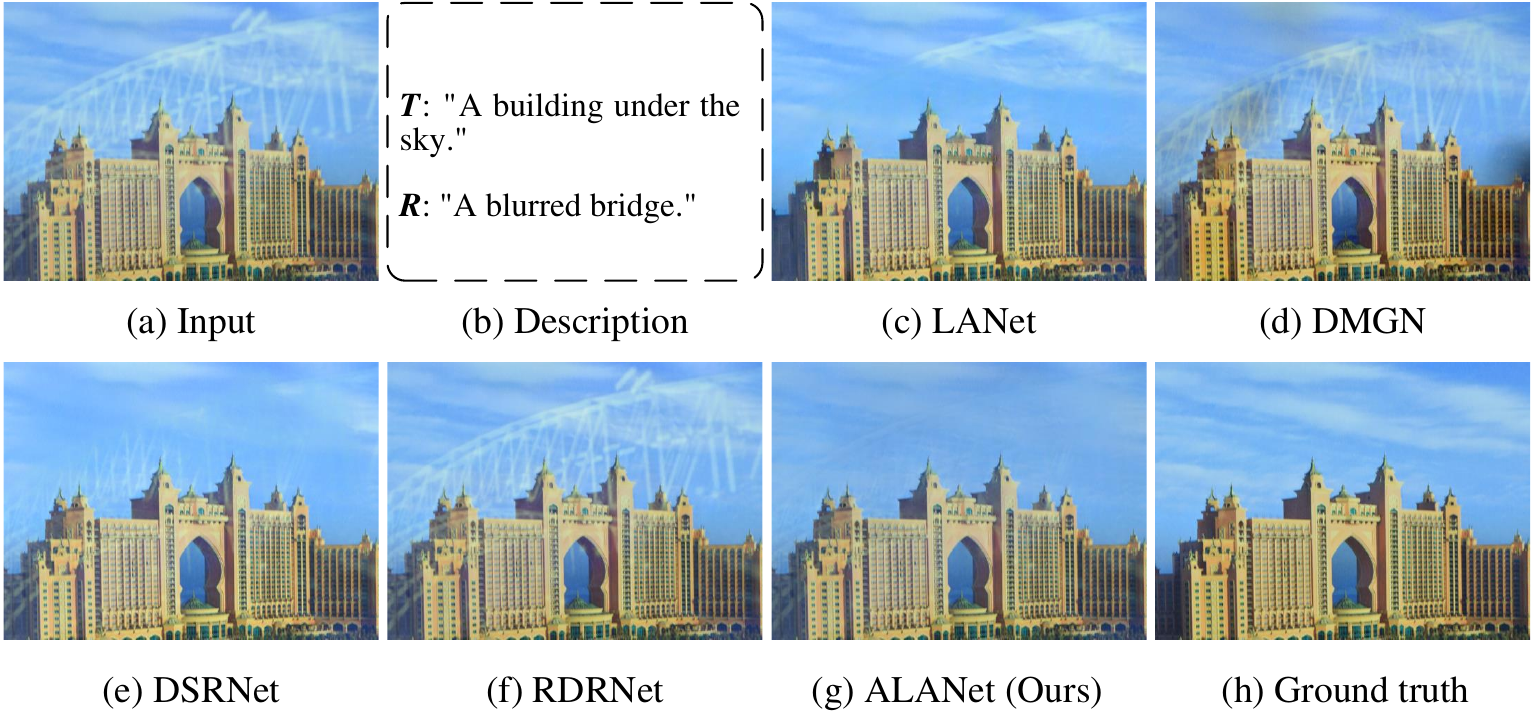} 
	\caption{Qualitative comparison with SOTA methods on the Postcard dataset.}
	\label{figv3}
\end{figure*}

\begin{figure*}[t]
	\centering
	\includegraphics[width=0.9\textwidth]{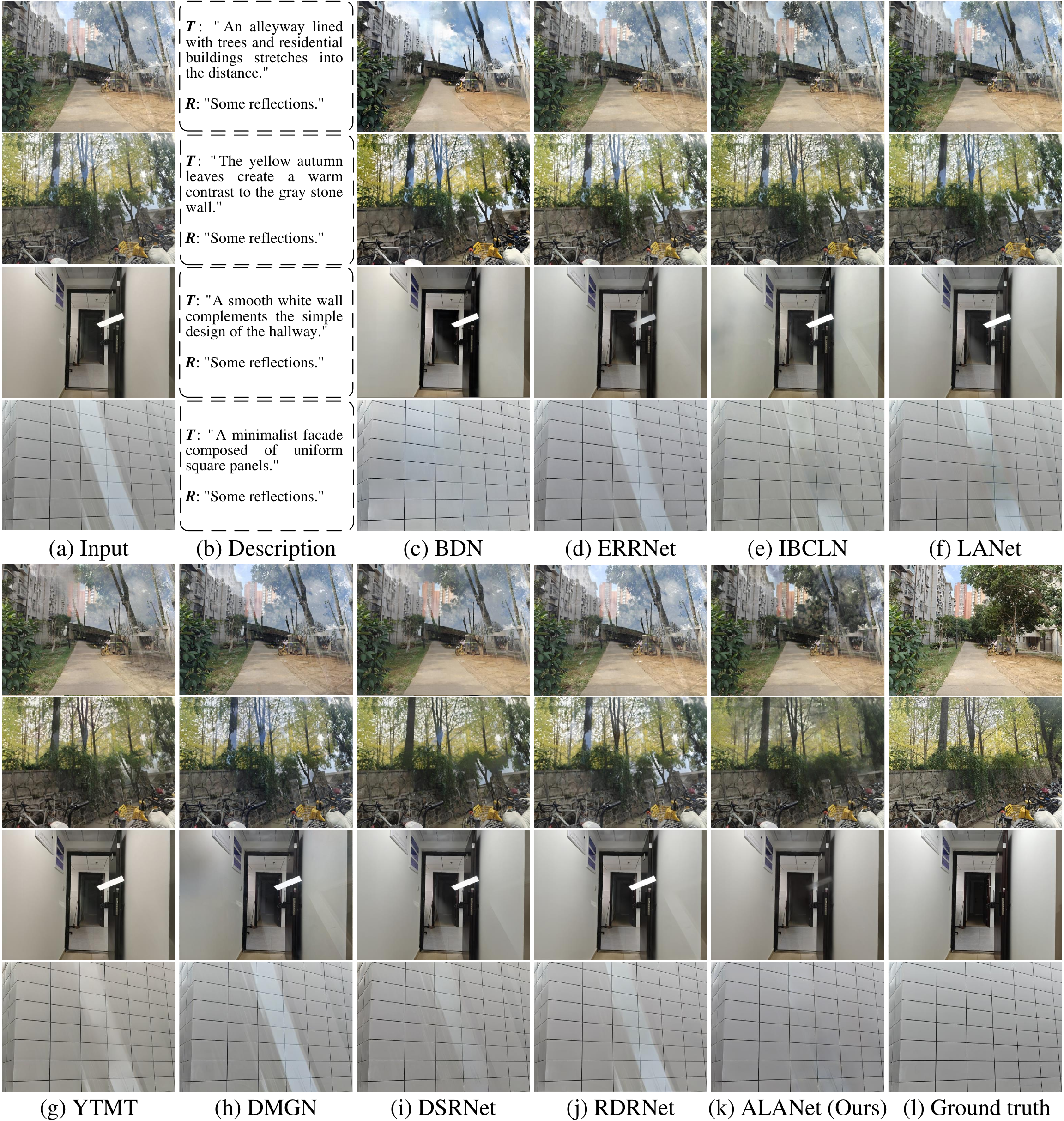} 
	\caption{Qualitative comparison with SOTA methods on the CRLAV dataset.}
	\label{figv4}
\end{figure*}

\begin{figure*}[t]
	\centering
	\includegraphics[width=0.9\textwidth]{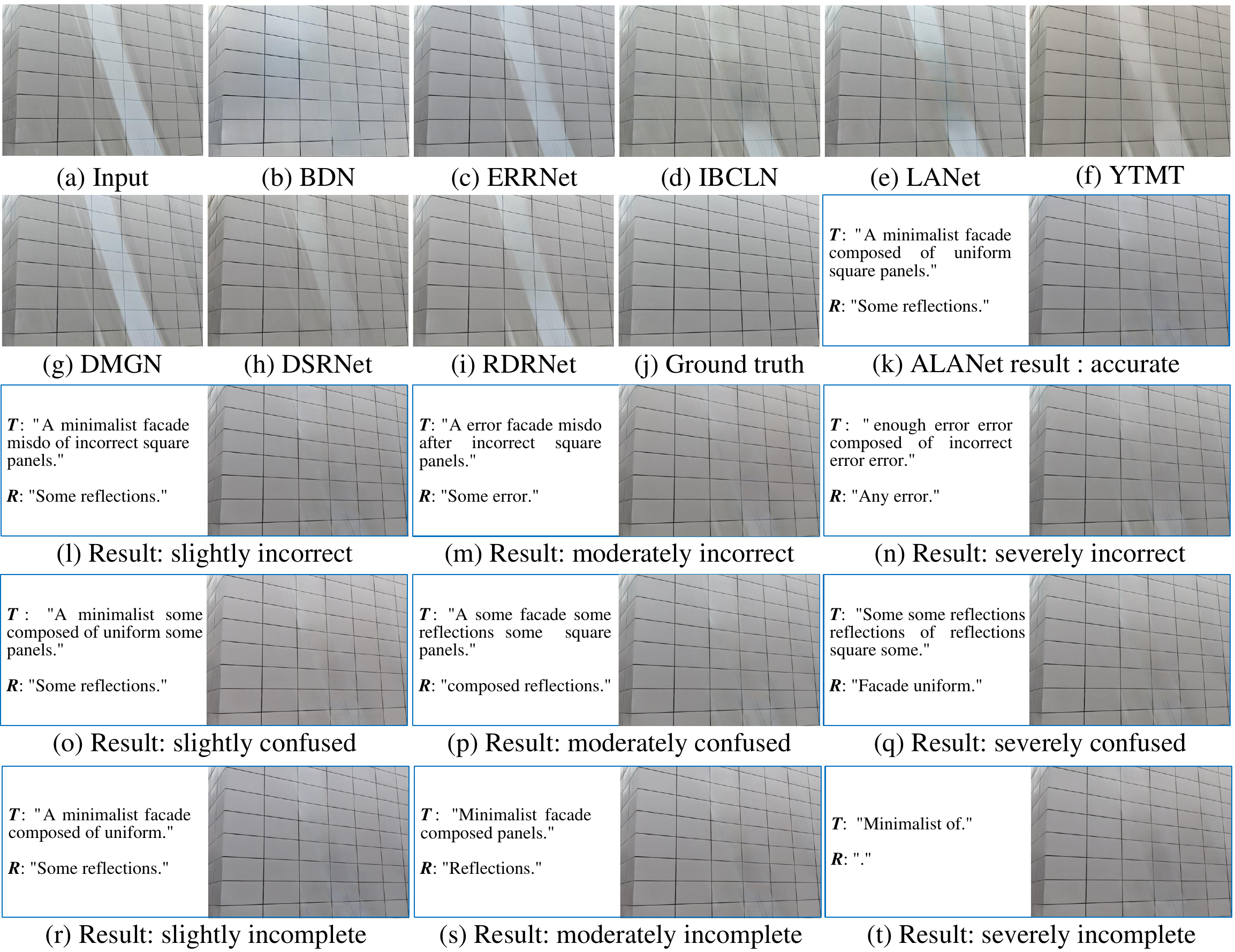} 
	\caption{Qualitative comparison of ALANet (under descriptions with varying language accuracies) and SOTA methods on the CRLAV dataset.}
	\label{compare2_varyinglanguageaccuracies}
\end{figure*}

\begin{figure*}[t]
	\centering
	\includegraphics[width=0.9\textwidth]{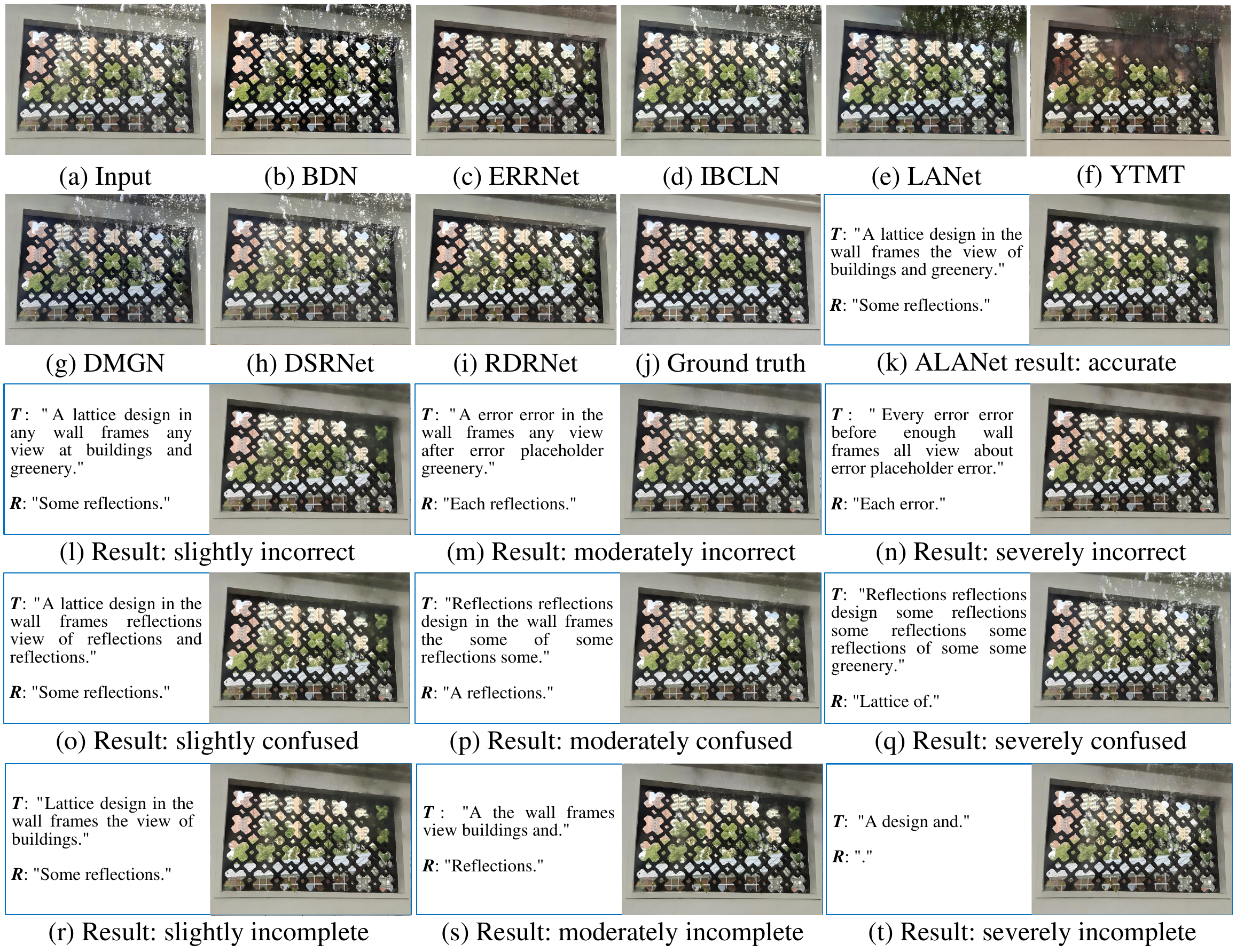} 
	\caption{Qualitative comparison of ALANet (under descriptions with varying language accuracies) and SOTA methods on the CRLAV dataset.}
	\label{compare1_varyinglanguageaccuracies}
\end{figure*}


\begin{figure*}[t]
	\centering
	\includegraphics[width=0.9\textwidth]{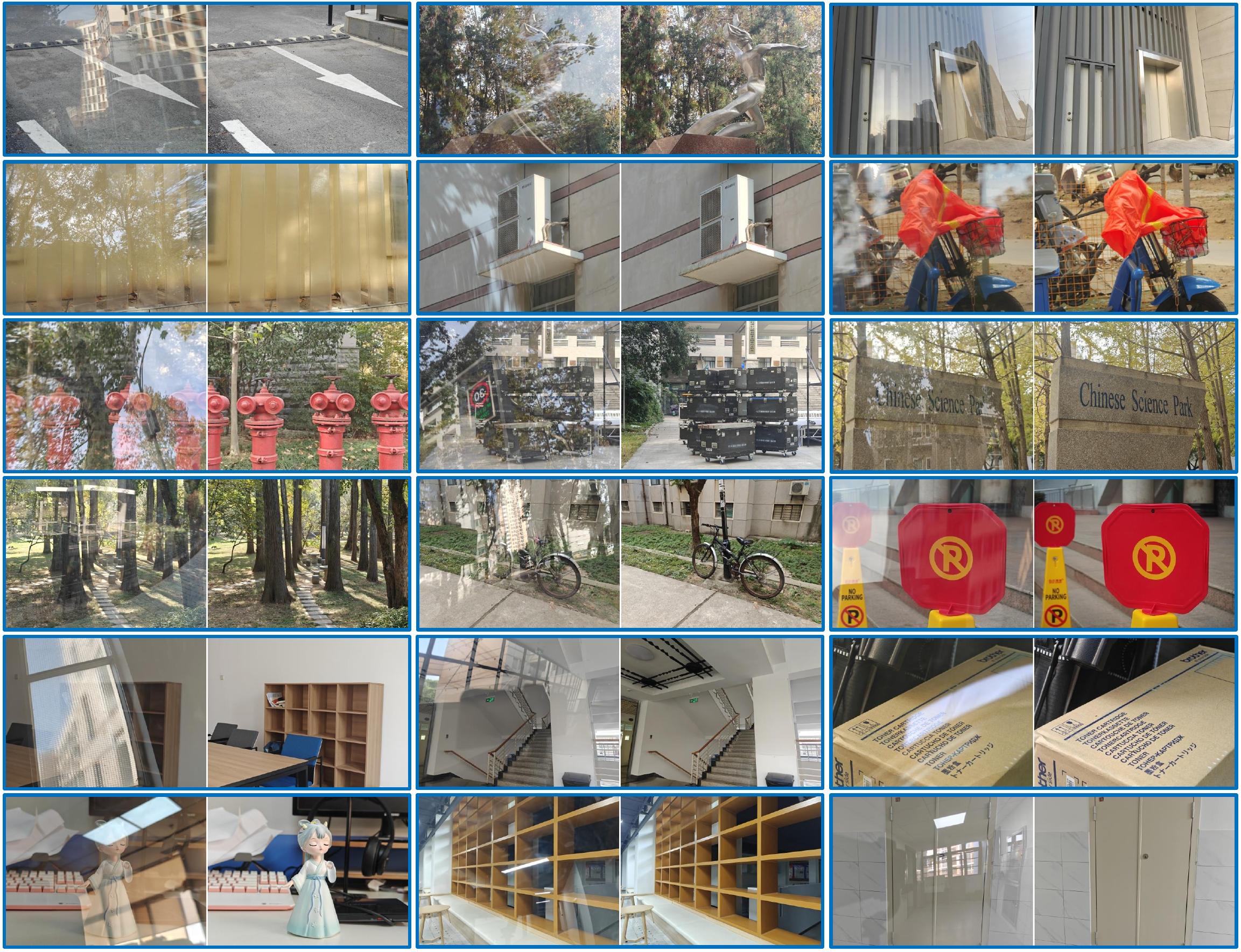} 
	\caption{Examples of images in the CRLAV dataset. Left: Reflected image; Right: Ground truth.}
	\label{figv6}
\end{figure*}

\begin{figure*}[t]
	\centering
	\includegraphics[width=0.9\textwidth]{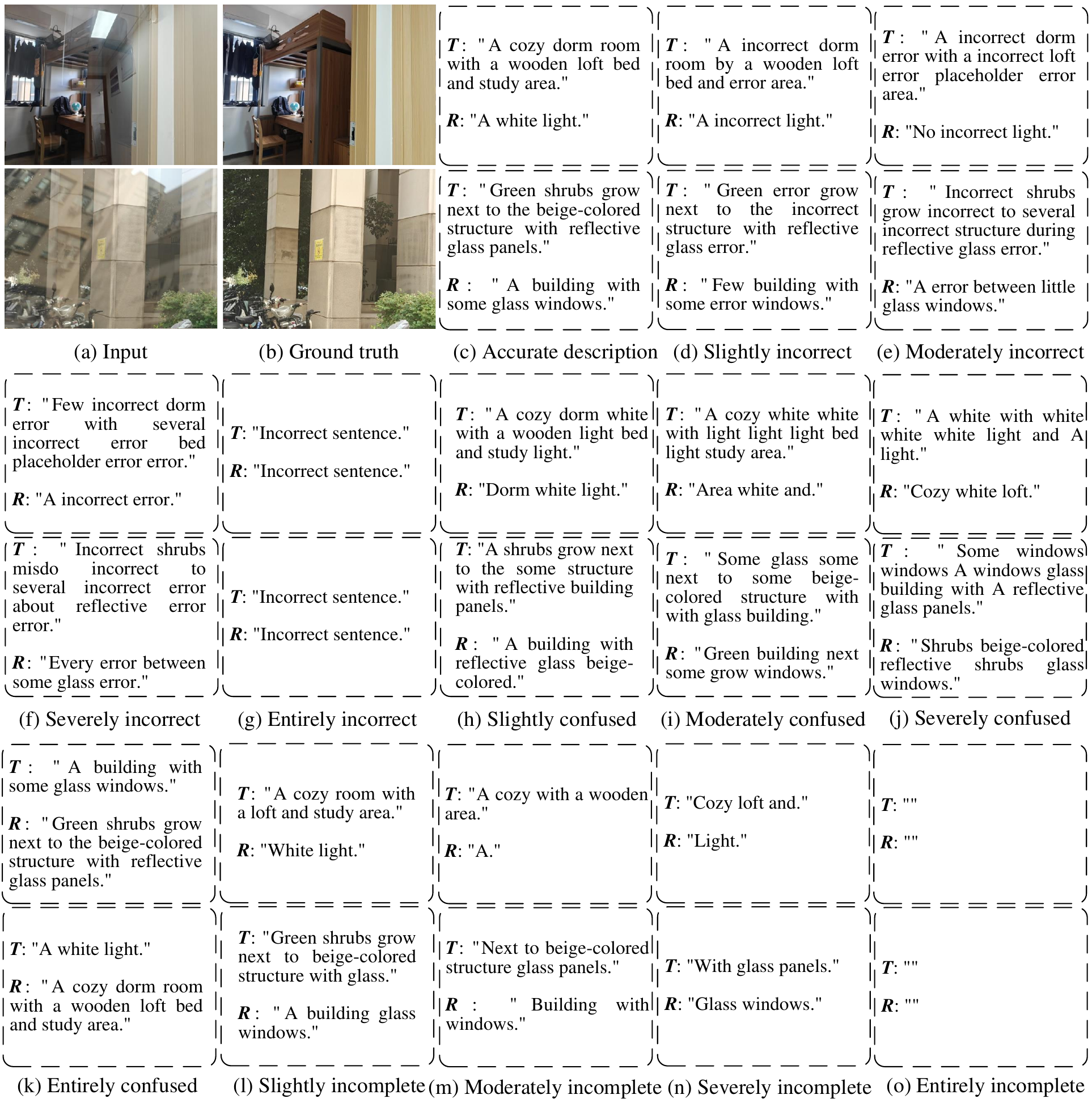} 
	\caption{Examples of images in the CRLAV dataset and their annotations with varying language accuracies.}
	\label{dataset_butongzhunquedu}
\end{figure*}

\bigskip
\end{document}